\documentclass[lettersize,journal]{IEEEtran}
\usepackage{amsmath,amsfonts}
\usepackage{algorithmic}
\usepackage{algorithm}
\usepackage{array}
\usepackage[caption=false,font=normalsize,labelfont=sf,textfont=sf]{subfig}
\usepackage{textcomp}
\usepackage{stfloats}
\usepackage{url}
\usepackage{verbatim}
\usepackage{graphicx}
\usepackage{etoolbox}
\usepackage{cite}
\usepackage{booktabs}
\usepackage{xcolor,colortbl}
\usepackage{amssymb}
\usepackage{wrapfig}
\usepackage{caption}
\usepackage{multirow} 
\usepackage{cuted}
\usepackage[pagebackref,breaklinks,colorlinks]{hyperref}
\definecolor{yellow}{rgb}{1, 1, 0.7}
\definecolor{orange}{rgb}{1, 0.85, 0.7}
\definecolor{cellred}{rgb}{1, 0.7, 0.7}
\definecolor{red}{rgb}{0, 0, 0}
\begin{document}
\title{VRP-UDF: Towards Unbiased Learning of Unsigned Distance Functions from Multi-view Images with Volume Rendering Priors}

\author{Wenyuan Zhang, Chunsheng Wang, Kanle Shi, Yu-Shen Liu~\IEEEmembership{Member,~IEEE,}
        Zhizhong Han% <-this % stops a space
\IEEEcompsocitemizethanks{\IEEEcompsocthanksitem Wenyuan Zhang and Yu-Shen Liu are with the School of Software, Tsinghua University, Beijing, China. E-mail: zhangwen21@mails.tsinghua.edu.cn, liuyushen@tsinghua.edu.cn.
\IEEEcompsocthanksitem Chunsheng Wang is with China Telecom Wanwei Information Technology Co., Ltd., Gansu, China. E-mail: v-chenxin.gs@chinatelecom.cn.
\IEEEcompsocthanksitem Kanle Shi is with Kuaishou Technology, Beijing, China. E-mail: shikanle@kuaishou.com.
\IEEEcompsocthanksitem Zhizhong Han is with the Department of Computer Science, Wayne State University, USA. E-mail: h312h@wayne.edu.
}% <-this % stops an unwanted space
\thanks{The corresponding author is Yu-Shen Liu. This work was partially supported by Deep Earth Probe and Mineral Resources Exploration -- National Science and Technology Major Project (2024ZD1003405), and the National Natural Science Foundation of China (62272263), and in part by Kuaishou. Project page: \url{https://wen-yuan-zhang.github.io/VolumeRenderingPriors}.}}

\markboth{Journal of \LaTeX\ Class Files,~Vol.~14, No.~8, August~2021}%
{Shell \MakeLowercase{\textit{et al.}}: A Sample Article Using IEEEtran.cls for IEEE Journals}

\maketitle

\begin{strip}
    \vspace{-1.5in}
    \centering
    \includegraphics[width=\textwidth]{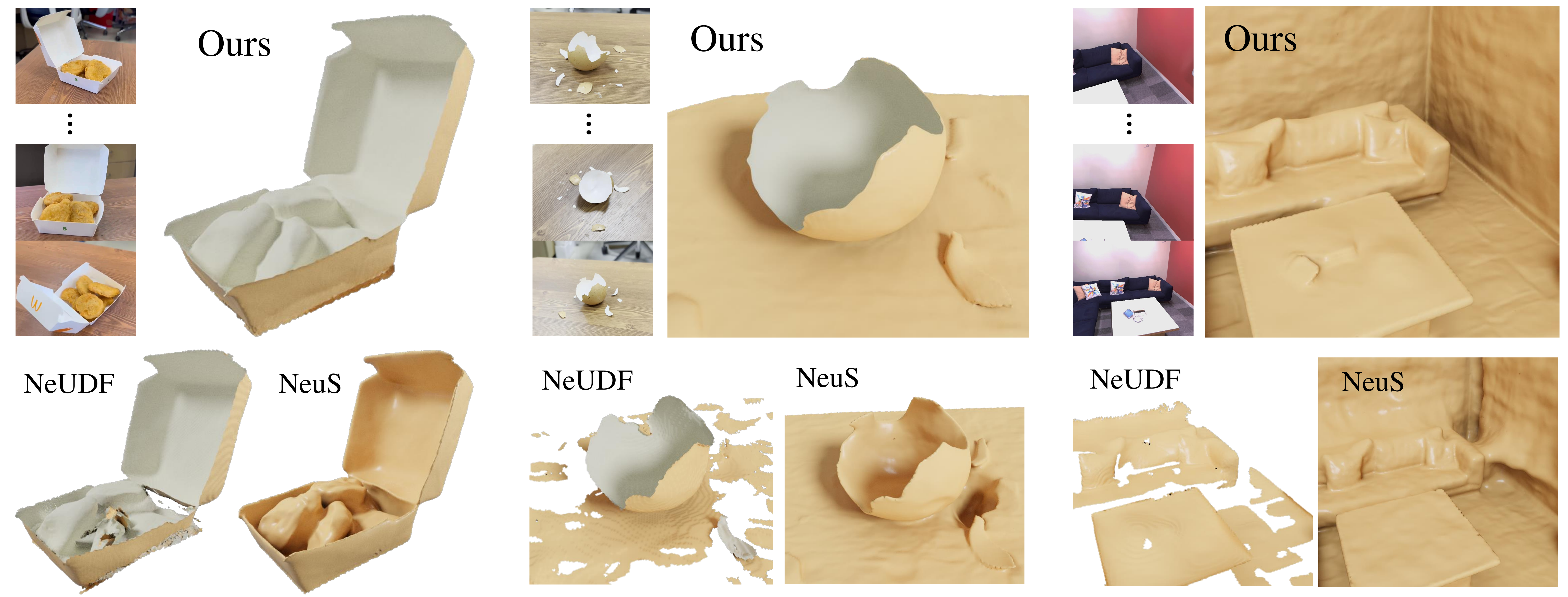}
    \captionof{figure}{We highlight our multi-view reconstruction results from UDFs learned on real-captured open surface scenes and indoor scenes. The two sides of a surface are colored in white and beige, respectively. Comparing with NeuS~\cite{wang2021neus} and the state-of-the-art UDF reconstruction method NeUDF~\cite{liu2023neudf}, our method does not produce artifacts and recovers more accurate and smoother geometries on both open and closed surfaces.}
    \label{fig:teaser}
\end{strip}

\begin{abstract}
Unsigned distance functions (UDFs) have been a vital representation for open surfaces. With different differentiable renderers, current methods are able to train neural networks to infer a UDF by minimizing the rendering errors with the UDF to the multi-view ground truth. However, these differentiable renderers are mainly handcrafted, which makes them either biased on ray-surface intersections, or sensitive to unsigned distance outliers, or not scalable to large scenes. To resolve these issues, we present a novel differentiable renderer to infer UDFs more accurately. Instead of using handcrafted equations, our differentiable renderer is a neural network which is pre-trained in a data-driven manner. It learns how to render unsigned distances into depth images, leading to a prior knowledge, dubbed volume rendering priors. To infer a UDF for an unseen scene from multiple RGB images, we generalize the learned volume rendering priors to map inferred unsigned distances in alpha blending for RGB image rendering. To reduce the bias of sampling in UDF inference, we utilize an auxiliary point sampling prior as an indicator of ray-surface intersection, and propose novel schemes towards more accurate and uniform sampling near the zero-level sets. We also propose a new strategy  that leverages our pretrained volume rendering prior to serve as a general surface refiner, which can be integrated with various Gaussian reconstruction methods to optimize the Gaussian distributions and refine geometric details. Our results show that the learned volume rendering prior is unbiased, robust, scalable, 3D aware, and more importantly, easy to learn. Further experiments show that the volume rendering prior is also a general strategy to enhance other neural implicit representations such as signed distance function and occupancy. We evaluate our method on both widely used benchmarks and real scenes, and report superior performance over the state-of-the-art methods. 
\end{abstract}


\section{Introduction}
\label{sec:intro}

\IEEEPARstart{N}{eural} implicit representations have become a dominant representation in 3D computer vision. Using coordinate-based deep neural networks, a mapping from locations to attributes at these locations like geometry~\cite{takikawa2021nglod, park2019deepsdf}, color~\cite{corona2022lisa, oechsle2020learning}, and motion~\cite{geng2023learning} can be learned as an implicit representation. Signed distance function (SDF)~\cite{park2019deepsdf} and unsigned distance function (UDF)~\cite{chibane2020neural} are widely used implicit representations to represent either closed surfaces~\cite{li2024gridformer, chen2024neuraltps} or open surfaces~\cite{zhou2024capudf, guillard2022meshudf}. We can learn SDFs or UDFs from supervisions like ground truth signed or unsigned distances~\cite{chabra2020deep, park2019deepsdf}, 3D point clouds~\cite{ma2021neuralpull, wang2022hsdf, chen2023gridpull,zhou2024fastnoise2noise, ma2022pcp, ma2022onsurfacepriors,zhou2023levelset,xiang2022snowflake,wen2022pmpnet++} or multi-view images~\cite{yariv2021volsdf, long2023neuraludf, huang2024neusurf, zhang2023fast, zhou2023differentiable}. Compared to SDFs, it is a more challenging task to estimate a UDF due to the sign ambiguity and the boundary effect, especially under a multi-view setting.

\begin{figure}[t]
\centering
  \includegraphics[width=\linewidth]{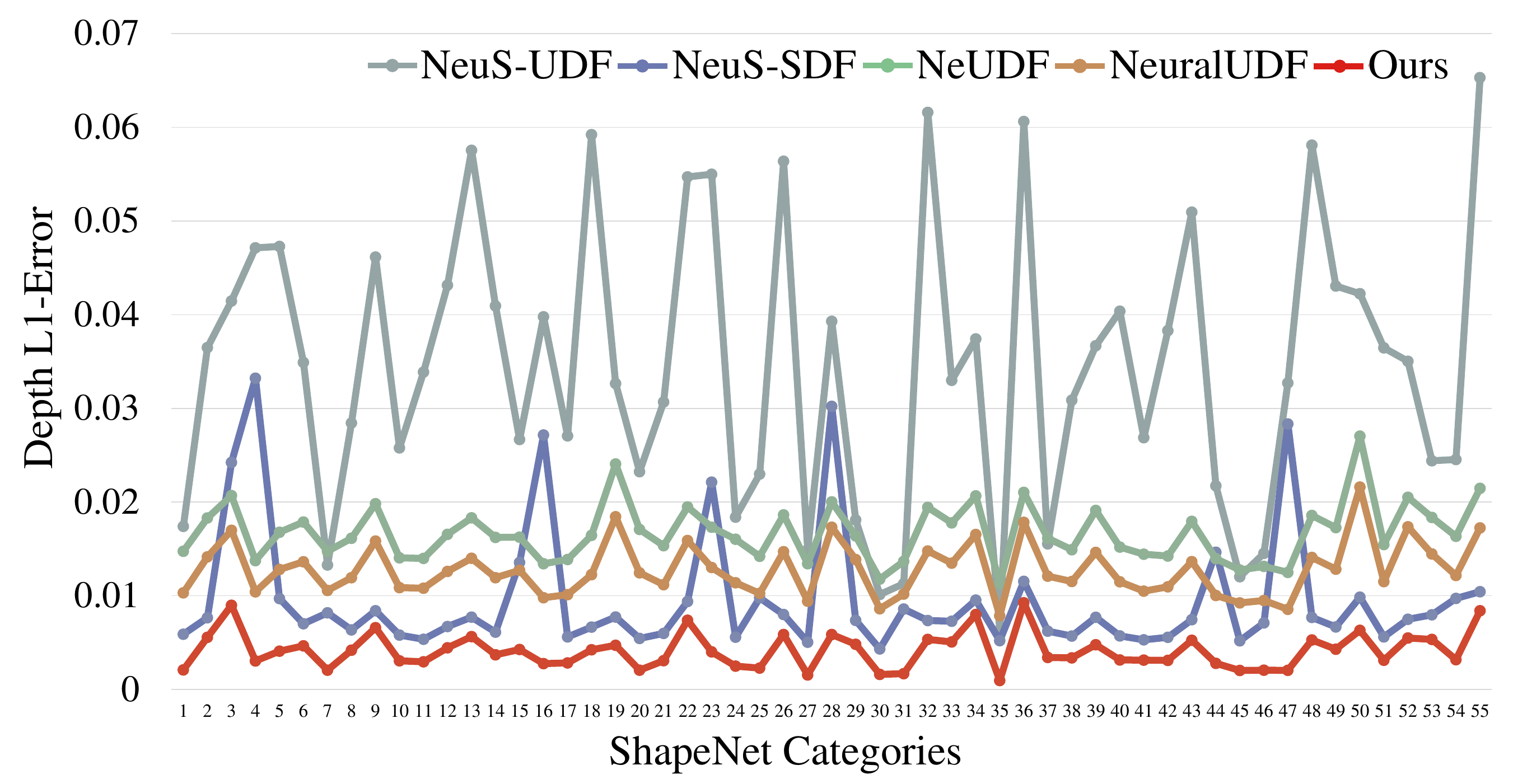}
    \caption{Statistics of depth L1-error for various differentiable renderers. Each data point represents the mean depth L1-error computed between 100 predicted and GT depth maps of a random object from each category of ShapeNet.}
    \label{fig:depth-error-curve}
\end{figure} 

\begin{figure}[t]
\centering
  \includegraphics[width=\linewidth]{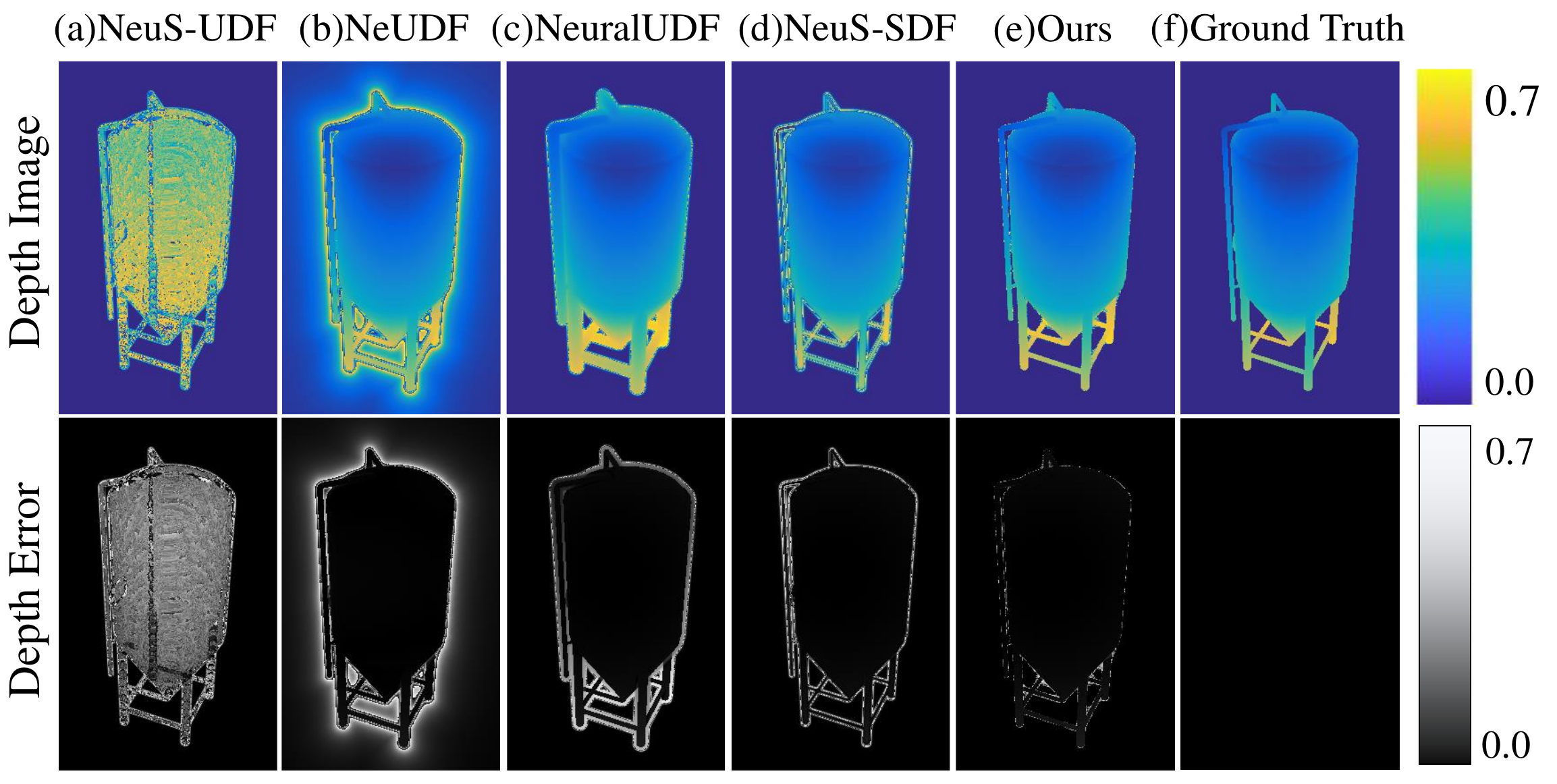}
    \caption{Comparisons of estimated depth images and depth error maps among different differentiable renderers on one shape from the category of ``tower'' in ShapeNet.}
    \label{fig:depth-error-visual}
\end{figure}

Recent methods~\cite{long2023neuraludf, liu2023neudf, deng20242sudf, meng2023neat} mainly infer UDFs from multi-view images through volume rendering. Using different differentiable renderers, they can render a UDF into RGB or depth images which can be directly supervised by the ground truth images. These differentiable renderers are mainly handcrafted equations which are either biased on ray-surface intersections, or sensitive to unsigned distance outliers, or not scalable to large scale scenes. These issues make them struggle with recovering accurate geometry.  Fig.~\ref{fig:depth-error-curve},~\ref{fig:depth-error-visual} detail these issues by comparing error maps on depth images rendered by different differentiable renderers. We render the ground truth UDF into depth images using different renderers from 100 different view angles, and report the average rendering error on each one of 55 shapes that are randomly sampled from each one of the 55 categories in ShapeNet~\cite{chang2015shapenet} in Fig.~\ref{fig:depth-error-curve}. Using the latest differentiable renderers from NeuS-UDF~\cite{wang2021neus} (using UDF as input to NeuS), NeUDF~\cite{liu2023neudf} and NeuralUDF~\cite{long2023neuraludf}, the rendered depth images and their error maps in Fig.~\ref{fig:depth-error-visual} (a)-(c) show that these issues cause large errors even using the ground truth UDF as inputs. Therefore, how to design better differentiable renderers for UDF inference from multi-view images is still a challenge.

To resolve these issues, we introduce a novel differentiable renderer for UDF inference from multi-view images through volume rendering. Instead of handcrafted equations used by the latest methods~\cite{long2023neuraludf, liu2023neudf, deng20242sudf, meng2023neat}, we employ a neural network to learn to become a differentiable renderer in a data-driven manner. Using UDFs and depth images obtained from meshes as ground truth, we train the neural network to map a set of unsigned distances at consecutive locations along a ray into weights for alpha blending, so that we can render depth images, and produce the rendering errors to the ground truth as a loss. We make the neural network observe different variations of unsigned distance fields during training, and learn the knowledge of volume rendering with unsigned distances by minimizing the rendering loss. The knowledge we call \textit{volume rendering prior} is highly generalizable to infer UDFs from multi-view RGB images in unobserved scenes. During testing, we use the pre-trained network as a differentiable renderer for alpha blending. It renders unsigned distances inferred by a UDF network into RGB images which can be supervised by the observed RGB images. Our results in Fig.~\ref{fig:depth-error-curve} and Fig.~\ref{fig:depth-error-visual} (e) show that we produce the smallest rendering errors among all differentiable renderers for UDFs, which is even more accurate than NeuS-SDF~\cite{wang2021neus} (rendering with ground truth SDF) in Fig.~\ref{fig:depth-error-visual} (d). 

We originally presented our method at ECCV2024~\cite{zhang2024learning}, and then extend our method in more applications with novel strategies of learning volume rendering priors and unbiased sampling. Specifically, we extend the architecture of the prior network from a single window size to multiple window sizes with multi-resolution feature fusion. This design allows the network to capture detailed features at different scales according to the UDF distributions within the neighborhood for alpha blending.

We also propose two strategies to mitigate the bias in the hierarchical sampling of UDFs. First, we apply an auxiliary pre-trained point sampling prior as a ray-surface intersection indicator. The output of the indicator is used to refine the sampling densities along the ray, preventing erroneous sampling near the surfaces. Second, we delve into the inverse transform sampling process and introduce a regularization term to promote the uniform distribution of upsampled points. These approaches significantly reduce the bias and enhance the robustness of the UDF hierarchical sampling. 

We additionally integrate our pretrained volume rendering priors into Gaussian reconstruction methods to serve as a general surface refiner. Taking advantage of dense queries around the UDF zero-level set, we push the Gaussian centers toward the direction of increasing accumulated rendering weights along the rays. This process facilitates the inference of accurate surfaces from sparse Gaussian distributions and significantly improves the reconstruction quality of the baseline methods.

Extensive experiments in our evaluations show that the learned volume rendering priors are unbiased, robust, scalable, 3D aware, and more importantly, easy to learn. We conduct evaluations in both widely used benchmarks and real scenes, and report superior performance over the state-of-the-art methods. We also generalize the volume rendering prior to other neural implicit representations, including SDF and occupancy, to showcase the generality and effectiveness of our volume rendering priors. Our contributions are listed below,

\begin{itemize}
    \item We introduce volume rendering priors to infer UDFs from multi-view images. Our prior can be learned in a data-driven manner, which provides a novel perspective to recover geometry with prior knowledge through volume rendering.
    \item We propose a novel deep neural network and learning scheme, and report extensive analysis to learn an unbiased differentiable renderer for UDFs with robustness, scalability, and 3D awareness.
    \item We propose two novel strategies to reduce the bias of hierarchical sampling by leveraging the volume rendering prior and optimizing the inverse sampling process.
    \item We extend our approach to Gaussian Splatting and propose a general strategy that consistently improves the performance of various Gaussian based reconstructions.
    \item We report the state-of-the-art reconstruction accuracy from UDFs inferred from multi-view images on widely used benchmarks and real image sets.
\end{itemize}

\section{Related Work}
\label{sec:related work}
\subsection{Multi-view 3D reconstruction}
Multi-view 3D reconstruction aims to reconstruct 3D shapes from calibrated images captured from overlapping viewpoints. The key idea is to leverage the consistency of features across different views to infer the geometry. MVSNet~\cite{yao2018mvsnet} is the first to introduce the deep learning-based idea into traditional MVS methods. Following studies explore the potential of MVSNet in different aspects, such as training speed~\cite{ weilharter2021highres}, memory consumption~\cite{yan2020dense} and network structure~\cite{ding2022transmvsnet}. These techniques produce depth maps or 3D point clouds. To obtain meshes as final 3D representations, additional procedures such as TSDF-fusion~\cite{curless1996volumetric} or classic surface reconstruction~\cite{kazhdan2013screened} methods are used, which is complex and not intuitive.

\subsection{Learning SDFs from Multi-view Images} 
Instead of 3D point clouds estimated by MVS methods, recent methods~\cite{wang2021neus, yariv2021volsdf} directly estimate SDFs through volume rendering from multi-view images for continuous surface representations. The widely used strategy is to render the estimated SDF into RGB images~\cite{darmon2022neuralwarp,jiang2025sensing} or depth images~\cite{yu2022monosdf, wang2022gosurf, azinovic2022neuralrgbd} which can be supervised by the ground truth images. The key to make the whole procedure differentiable is various differentiable renderers~\cite{wang2021neus, azinovic2022neuralrgbd, niemeyer2020differentiable} which transform signed distances into weights for alpha blending during rendering. Some methods modify the rendering equations to use more 2D supervisions like normal maps~\cite{wang2022neuris,zhang2025nerfprior}, detected planes~\cite{zhou2024planar, han2025sparserecon}, and segmentation maps~\cite{kong2023vmap,zhang2025monoinstance} to pursue higher reconstruction efficiency. Similarly, some methods~\cite{zhu2022nice-slam, zhu2024nicer} jointly estimate camera poses and geometry in the context of SLAM. Recent efforts have attempted to recover continuous SDF fields from discrete 3D Gaussians~\cite{kerbl20233d}. These methods typically combine 3DGS with neural volume rendering~\cite{chen2023neusg,yu2024gsdf,li2025vags} or differentiable pulling operations~\cite{zhang2024gspull, noda2024multipull,li2025gaussianudf}, accompanied by a series of regularization constraints~\cite{guedon2024sugar, huang20242dgs,chen2024pgsr,zhang2025materialrefgs}. However, the SDFs that these methods aim to learn are only for closed surfaces, which is limited to represent open surfaces.

\subsection{Learning UDFs from Multi-view Images} 
Different from SDFs, UDFs~\cite{chibane2020neural,zhou2024udiff,zhou2024capudf,li2025ifiltering,zhang2025gap,li2025gaussianudf} are able to represent open surfaces. Recent methods~\cite{long2023neuraludf, meng2023neat, deng20242sudf, liu2023neudf} design different differentiable renderers to learn UDFs through multi-view images. NeuralUDF~\cite{long2023neuraludf} predicts the first intersection along a ray and flips the UDFs behind this point to use the differentiable renderer of NeuS~\cite{wang2021neus}. NeUDF~\cite{liu2023neudf} proposes an inverse proportional function mapping UDF to rendering weights. NeAT~\cite{meng2023neat} learns an additional validity probability net to predict the regions with open structures, while 2S-UDF~\cite{deng20242sudf} proposes a bell-shaped weight function that maps UDF to density in a two-stage manner. The differentiable renderers introduced by these methods mainly get formulated into handcrafted equations which are biased on ray-surface intersections, sensitive to unsigned distance outliers, and not 3D aware. We resolve this issue by introducing a learning-based differentiable renderer which learns and generalizes a volume rendering prior for robustness and scalability. The ideas of learnable neural rendering frameworks are also introduced in~\cite{chang2023pointersect, arandjelovic2021nerfindetail, kurz2022adanerf}.

\noindent \textbf{Hierarchical Volume Sampling.} NeRF~\cite{mildenhall2020nerf} proposes a two-stage hierarchical volume sampling (HVS) strategy to increase the rendering efficiency. Subsequent works optimize the positions of sampling rays and sampling points by employing more comprehensive theoretical models~\cite{li2024l0sampler}, introducing auxiliary networks~\cite{arandjelovic2021nerfindetail, kurz2022adanerf}, incorporating rendering results~\cite{zhang2023fast, sun2024efficient}, and improving ray representations~\cite{barron2022mipnerf360}. However, these methods are mostly tailored for density-based volume rendering~\cite{mildenhall2020nerf, barron2022mipnerf360} and are not applicable to neural implicit function-based volume rendering processes~\cite{wang2021neus, long2023neuraludf}. NeuS~\cite{wang2021neus} proposes an unbiased sampling strategy for SDF-based volume rendering, but this design becomes more complex when applied to UDFs, as UDFs lack an explicit distinction of zero-level set. Building upon the sampling approach of NeUDF~\cite{liu2023neudf}, we introduce an auxiliary point sampling prior and optimize the inverse transform sampling to significantly mitigate the bias in the hierarchical sampling process.

%-----------------------------------------------------
%-----------------------------------------------------

\begin{figure*}[t]
  \centering
  \includegraphics[width=\linewidth]{Figures/Overview-small.pdf}
  \caption{Overview of our method. In the training phase, our volume rendering prior takes sliding windows of GT UDFs from training meshes as input, and outputs opaque densities for alpha blending. The parameters are optimized by the error between rendered depth and ground truth depth maps. During the testing phase, we freeze the volume rendering prior and use ground truth multi-view RGB images to optimize a randomly initialized UDF field.}
  \vspace{-0.1cm}
  \label{fig:overview}
\end{figure*}

\section{Method}

\subsection{Overview}

Given a set of $J$ images $\{I_j\}_{j=1}^{J}$, we aim to infer a UDF $f_u$ which predicts an unsigned distance $u$ for an arbitrary 3D query $q$. We formulate the UDF as $u=f_u(q)$. With the learned $f_u$, we can extract the zero level set of $f_u$ as a surface using algorithms similar to the marching cubes~\cite{guillard2022meshudf, zhou2024capudf}.

We employ a neural network to learn $f_u$ by minimizing rendering errors to the ground truth. We shoot rays from each view $I_j$, sample queries $q$ along each ray, and get unsigned distance prediction $u$ from $f_u$ to calculate weights $w$ for alpha blending in volume rendering. At the same time, we train a color function $c$ which predicts the color at these queries $q$ as $c=f_c(q)$. The accumulation of $c$ with weights $w$ along the ray produces a color at the pixel.

Current differentiable renderers~\cite{long2023neuraludf, deng20242sudf, liu2023neudf, meng2023neat} transform $u$ into $w$ using handcrafted equations. Instead, we train a neural network to approximate this function $f_w$ in a data-driven manner, as illustrated in Fig.~\ref{fig:overview}. During training, we push $f_w$ to produce ideal weights for rendering depth images that are as similar to the supervision $\{D_a^h\}_{a=1}^A$ from the $h$-th shape as possible using the ground truth UDF, and more importantly, get used to various variations of unsigned distances along a ray. During testing, we use this volume rendering prior with fixed parameters $\theta_w$ of $f_w$. We leverage $f_w$ to estimate an $f_u$ from multi-view RGB images $\{I_j\}_{j=1}^J$ of an unseen scene by minimizing rendering errors of RGB color through volume rendering.

\begin{figure}[t]
  \centering
  \includegraphics[width=\linewidth]{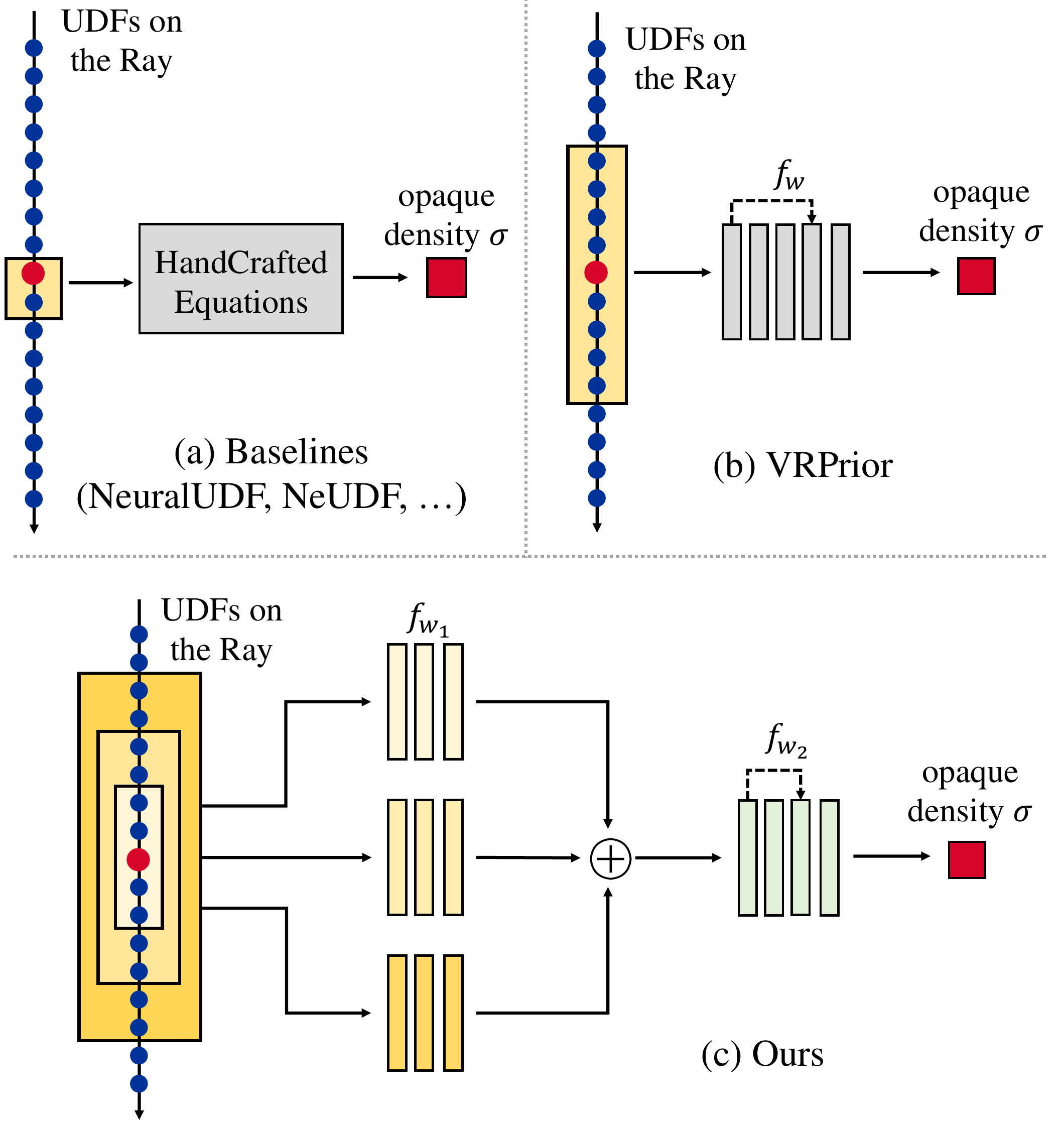}
  \caption{Comparisons of different differentiable renderer structures. Existing methods use handcrafted equations to convert UDFs to opaque density. We extend the single-resolution MLP of VRPrior to multi-resolution MLPs, which further enhance the robustness and 3D awareness of volume rendering priors in neighborhood.}
  \label{fig:method-mlp-structure}
\end{figure}

\subsection{Volume Rendering for UDFs}
\label{sec3.2}
We render a UDF function $f_u$ with a color function $f_c$ into either RGB $I'$ or depth $D'$ images to compare with the RGB supervision $\{I_j\}_{j=1}^J$ or depth supervision $\{D_j\}$. 
% Note that we do not use depth supervision $\{D_j\}$ during the UDF inference, but we include a depth supervision here to make the UDF rendering with learned priors self-contained.
From each posed view $I_j$, we sample some pixels and shoot rays starting at each pixel. Taking a ray $V_k$ from view $I_j$ for example, $V_k$ starts from the camera origin $o$ and points to a direction $r$. We hierarchically sample $N$ points along the ray $V_k$, where each point is sampled at $q_n=o+d_n*r$ and $d_n$ corresponds to the depth value of $q_n$ on the ray. We can transform unsigned distances $f_u(q_n)$ into weights $w_n$ which is used for color or depth accumulation along the ray $V_k$ in volume rendering,
\begin{equation}
\label{eq:volumerendering}
\begin{split}
\sigma_n&=f_w(\{q_m\}_{m=1}^M,\{f_u(q_m)\}_{m=1}^M), \\
w_n&=\sigma_n\times \prod\nolimits_{n'=1}^{n-1}(1-\sigma_{n'}) \\
I(k)'&=\sum\nolimits_{n'=1}^{N} w_{n'}\times f_c(q_{n'}), \\
D(k)'&=\sum\nolimits_{n'=1}^{N} w_{n'}\times d_{n'},
\end{split}
\end{equation}
\begin{figure}[t]
  \centering
  \includegraphics[width=\linewidth]{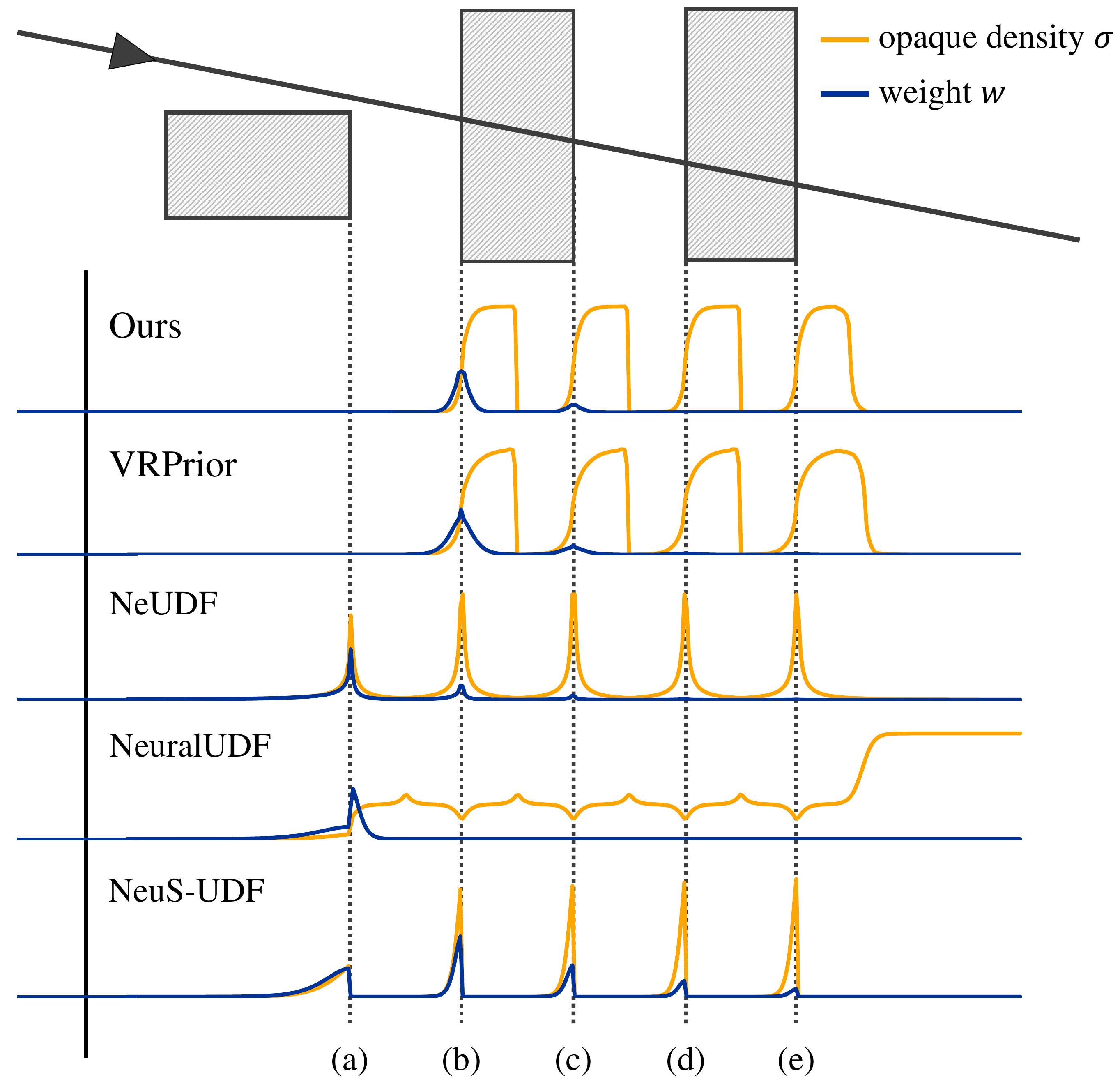}
  \caption{Distribution of opaque densities and accumulated weights calculated by different baselines and predicted by our volume rendering priors. Our method is 3D aware and robust to unsigned distance changes at near-surface points while deriving unbiased volume rendering weights.}
  \label{fig:volumerendering-compare}
\end{figure}

\noindent where $q_m$ is one of $M$ nearest neighbors of $q_n$ along a ray, and $\sigma_n$ is the opaque density that can be interpreted as the differential probability of a ray terminating at an infinitesimal particle at the location $q_n$. The latest methods model the weighting function $f_w$ in handcrafted ways with $M=2$ neighbors around $q_n$ on the same ray. For instance, NeUDF~\cite{liu2023neudf} introduced an inverse proportional function to calculate opaque density from two adjacent queries, while NeuralUDF~\cite{long2023neuraludf} used the same two queries to model both occlusion probability and opaque densities. 

Although these differentiable renderers can render UDFs into images, they are usually biased at intersection of ray and surface and produce rendering errors on the boundaries on depth images, as shown by error maps in Fig.~\ref{fig:depth-error-visual} (b) and (c). These errors indicate that these handcrafted equations can not render correct depth even when using the ground truth unsigned distances as supervision. 

Why do these handcrafted equations produce large rendering errors? Our analysis shows that being not 3D aware plays a key role in producing these errors. These handcrafted equations merely use $M$=2 neighboring points to perceive the 3D structure when calculating the opaque density at query $q_n$, as shown in Fig.~\ref{fig:method-mlp-structure} (a). Such a small window makes these equations merely have a pretty small receptive field, which makes them become sensitive to unsigned distance changes, such as the weight decrease at queries sampled on a ray that is passing by an object. Moreover, to approximate some characteristics like unbiasedness and occlusion awareness, these equations are strictly handcrafted, which make them extremely hard to get extended to be more 3D aware by using more neighboring points as input. Another demerit comes from the fact that all rays need to use the same equation to model the opaque, which is not generalizable enough to cover various unsigned distance combinations.

Fig.~\ref{fig:volumerendering-compare} illustrates issues of current methods. When a ray is approaching an object, handcrafted equations struggle to produce a zero opaque density at the location where the ray merely passes by an object but not intersects with it. This is also the reason why these methods produce large rendering errors on the boundary in Fig.~\ref{fig:depth-error-visual}. To resolve these issues, we introduce to train a neural network to learn the weight function $f_w$ in a data-driven manner, which leads to a volume rendering prior. During training, the network observes huge amount of unsigned distance variations along rays, and learns how to map unsigned distances into weights for alpha blending.

\subsection{Learning Volume Rendering Priors}
\label{sec3.3}
Our data-driven strategy uses ground truth meshes $\{S_h\}_{h=1}^H$ to learn the function $f_w$. For each shape $S_h$, we calculate its ground truth UDF $f_{u_{gt}}^h$, render $A$=100 ground truth depth images $\{D^h_a\}_{a=1}^A$ from randomly sampled view angles around it, and push the neural network learning $f_w$ to render depth images $\{\tilde{D}^h_a\}_{a=1}^A$ to be as similar to $\{D^h_a\}_{a=1}^A$ as possible. During rendering, we leverage $f_{u_{gt}}^h$ to provide ground truth unsigned distances at query $q_n$, which leaves the function $f_w$ as the only learning target.

Specifically, along a ray $V_k$, we hierarchically sample $N$=128 queries $\{q_n\}$ to render a depth value through volume rendering using Eq.~\eqref{eq:volumerendering}. We employ a sampling strategy detailed in Sec.~\ref{sec3.5}. For each query $q_n$, we calculate its ground truth unsigned distance $u_n\!=\!f_{u_{gt}}^h(q_n)$ and the ground truth unsigned distances at its neighboring points $\{u_m'\!=\!f_{u_{gt}}^h(q_m')\}_{m=1}^M$. Besides, we also use the sampling interval $\delta_m$ between $q_m'$ and $q_{m+1}'$
as another clue. Therefore, we formulate the modeling of opaque density as,
\begin{equation}
\label{eq:opaque}
\sigma_n=f_w(\{\delta_m\}_{m=1}^M,\{f_{u_{gt}}^h(q_m')\}_{m=1}^M), q_m'\in NN(q_n).
\end{equation}
Instead of handcrafted equations, we use a neural network to model the function $f_w$. It will learn a volume rendering prior which is a prior knowledge of being a good renderer for UDFs. Obviously, it is more adaptive to different rays than handcrafted equations, and become more 3D aware with the flexibility of using a larger neighboring size.

\begin{figure}[t]
  \centering
  \includegraphics[width=\linewidth]{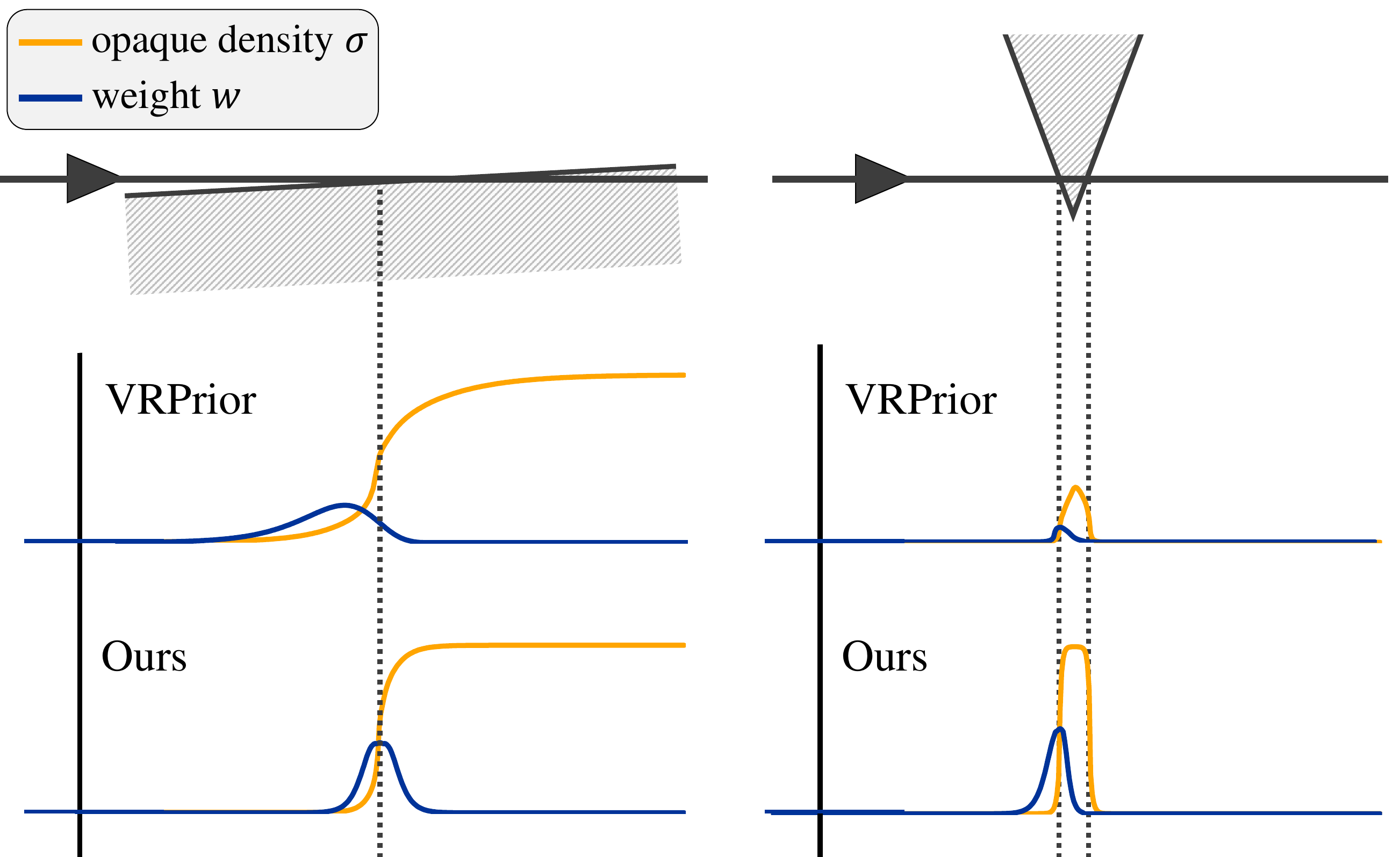}
  \vspace{-0.1cm}
  \caption{Comparison between VRPrior and our method. Our multi-resolution windows enable significantly more accurate transformations in handling different UDF variants, such as large near-surface regions or sharp surface corners.}
  \label{fig:volumerendering-compare-vrprior-ours}
\end{figure}

To achieve this, we first divide the neighboring points into three groups based on their distances to $q_n$, as illustrated in Fig.~\ref{fig:method-mlp-structure} (c), where each group includes $M_1$=10, $M_2$=20 and $M_3$=30 points. For each group, we concatenate all of the unsigned distances $\{u_m'\}$ and sampling intervals $\{\delta_m\}$, and feed them into three separate MLPs $f_{w_1}$ with non-shared parameters to extract three features. The features are then added together and fed into another MLP $f_{w_2}$ with skip connection to predict the opaque density $\sigma_n$ at $q_n$. Unlike our previous work~\cite{zhang2024learning} that uses a fixed single window size as shown in Fig.~\ref{fig:method-mlp-structure} (b), we extract features from neighborhoods with different window sizes and fuse them to produce the final result. We find that the proposed multi-resolution MLPs further enhance the generalization ability of our volume rendering priors when dealing with UDF distributions with different patterns. As highlighted in Fig.~\ref{fig:volumerendering-compare-vrprior-ours}, when encountering scenarios such as a ray slowly passing through a surface (left in the figure) or traversing a sharp object corner (right in the figure), VRPrior~\cite{zhang2024learning} shows degraded performance. In contrast, our method adaptively infers the opaque densities by aggregating features from multi-scale window sizes, ensuring greater flexibility and robustness. We train all involved networks parameterized by $\theta_w$ by minimizing the rendering errors on depth images,

\begin{equation}
\label{eq:opaquelearning}
\min_{\theta_w} \sum\nolimits_{h=1}^{H}\sum\nolimits_{a=1}^{A} ||D_a^h-\tilde{D}^h_a||_2^2 .
\end{equation}

Note that we do not employ RGB images in the learning of priors, because our goal at this stage is to learn a $f_w$ to map UDFs to opaque densities for alpha blending. The additional color network is unnecessary and would reduce the generalization ability of the learned prior to unseen cases. The improvements brought by our prior are shown in Fig.~\ref{fig:volumerendering-compare}. We can accurately predict opaque densities with 3D awareness at arbitrary locations and robustness to unsigned distance changes.

\begin{figure}[t]
  \centering
  \includegraphics[width=\linewidth]{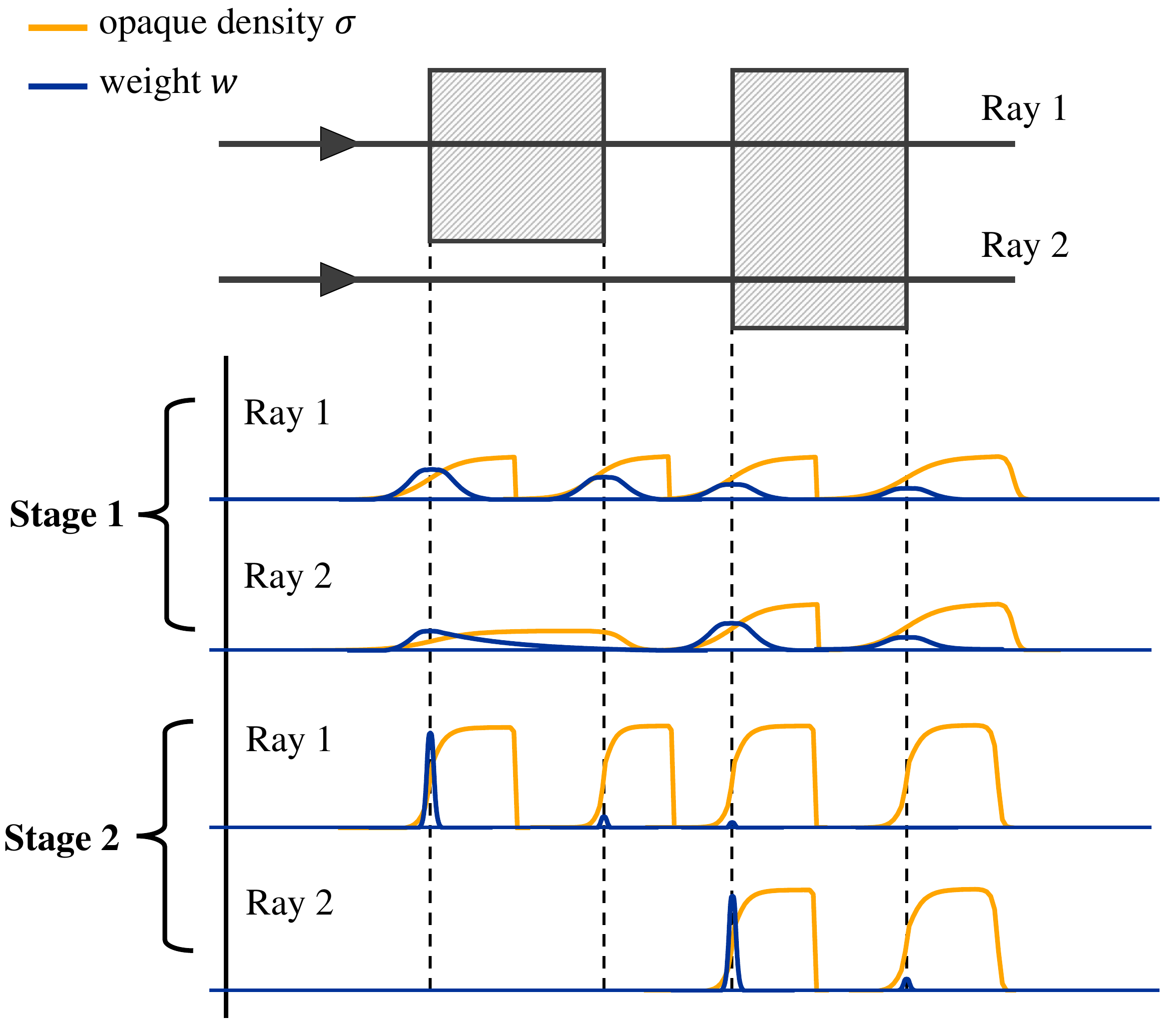}
  \vspace{-0.2cm}
  \caption{Visualization of opaque density and rendering weight distributions of stage 1 and stage 2. Stage 1 perceives a broad range of potential surface locations, while stage 2 concentrates the rendering weights on the first true surface.}
  \label{fig:revision-curve-twockpt}
\end{figure}

\subsection{Generalizing Volume Rendering Priors}
\label{sec3.4}
We use the volume rendering prior represented by $\theta_w$ of $f_w$ to estimate a UDF $f_u$ from a set of RGB images $\{I_j\}_{j=1}^J$ of an unseen scene. We learn $f_u$ by minimizing the rendering errors on RGB images. 

Specifically, for a ray $V_k$, we hierarchically sample $N$=128 queries $\{q_n\}$ to render RGB values through volume rendering using Eq.~\eqref{eq:volumerendering}. Similarly, we calculate the opaque density at each location $q_n$ using Eq.~\eqref{eq:opaque}, but using unsigned distances predicted by $f_u$ as $\sigma_n=f_w(\{\delta_m\}_{m=1}^M,\{f_u(q_m')_{m=1}^M\})$, not the ground truth ones when learning $f_w$, and keeping the parameters $\theta_w$ of $f_w$ fixed during the generalizing procedure. Although our model is trained on ground truth UDFs, it can effectively converge during generalization when initialized from a spherical UDF field. This is because our volume rendering prior learns a local mapping from UDFs to opaque densities, and is exposed to abundant local UDF patterns through multi-view optimization and random sampling during training. Under this setting, the multi-view photometric loss is back-propagated through volume rendering and $f_w$ to the entire UDF field, which drives the UDFs to gradually converge. Moreover, we use two neural networks to model the UDF $f_u$ and the color function $f_c$ parameterized by $\theta_u$ and $\theta_c$, respectively. We jointly learn $\theta_u$ and $\theta_c$ by minimizing the errors between rendered RGB images $\{\tilde{I}_j\}_{j=1}^J$ and the ground truth below,
\begin{equation}
\label{eq:colorloss}
\mathcal{L}_{rgb}=\sum\nolimits_{j=1}^{J}||I_j-\tilde{I}_j||.
\end{equation}
Our loss function is formulated with an additional Eikonal loss~\cite{yariv2020idr} $\mathcal{L}_{e}$ for regularization in the field,
\begin{equation}
\label{eq:totalloss}
\mathcal{L}= \mathcal{L}_{rgb}+\lambda \mathcal{L}_{e},
\end{equation}
\noindent where $\lambda$ is a balance weight and set to 0.1 following previous work~\cite{wang2021neus}. Using the learned parameters $\theta_u$, we use the method introduced in MeshUDF~\cite{guillard2022meshudf} to extract the zero-level set of $f_u$ as the surface.

To facilitate the smoothness learning of neural distance functions, existing methods typically introduce a learnable variance parameter to control the sharpness of the mapping functions~\cite{wang2021neus,yariv2021volsdf,liu2023neudf}. However, there is no explicit supervision that defines the correspondence between the variance and the UDF mapping functions. To address this issue, we propose a two-stage parameter strategy that employs two parameter sets of $f_w$ for the early and late stages of UDF inference. Specifically, the parameters of $f_w$ are saved twice during training, once in the middle and once at the end. The weight decay of the optimizer is reduced as training progresses. We observe that in the middle stage with a large weight decay, the network $f_w$ exhibits an underfitting behavior and produces a loose mapping that is less sensitive to UDF variations, which enables it to perceive a wide range of potential surface locations, as shown in ``stage 1'' of Fig.~\ref{fig:revision-curve-twockpt}. At the end of training with a small weight decay, the network learns sharp boundaries around the surfaces, resulting in rendering weights that are more concentrated on the zero-level set, as shown in ``stage 2'' of Fig.~\ref{fig:revision-curve-twockpt}. During the UDF inference, we firstly load the first parameters set as our prior, which recovers a coarse UDF representing the overall geometry. Then we load the second parameters set, which will refine the UDFs near the already determined surfaces, producing fine details and sharp edges.

\begin{figure}[t]
  \centering
  \includegraphics[width=\linewidth]{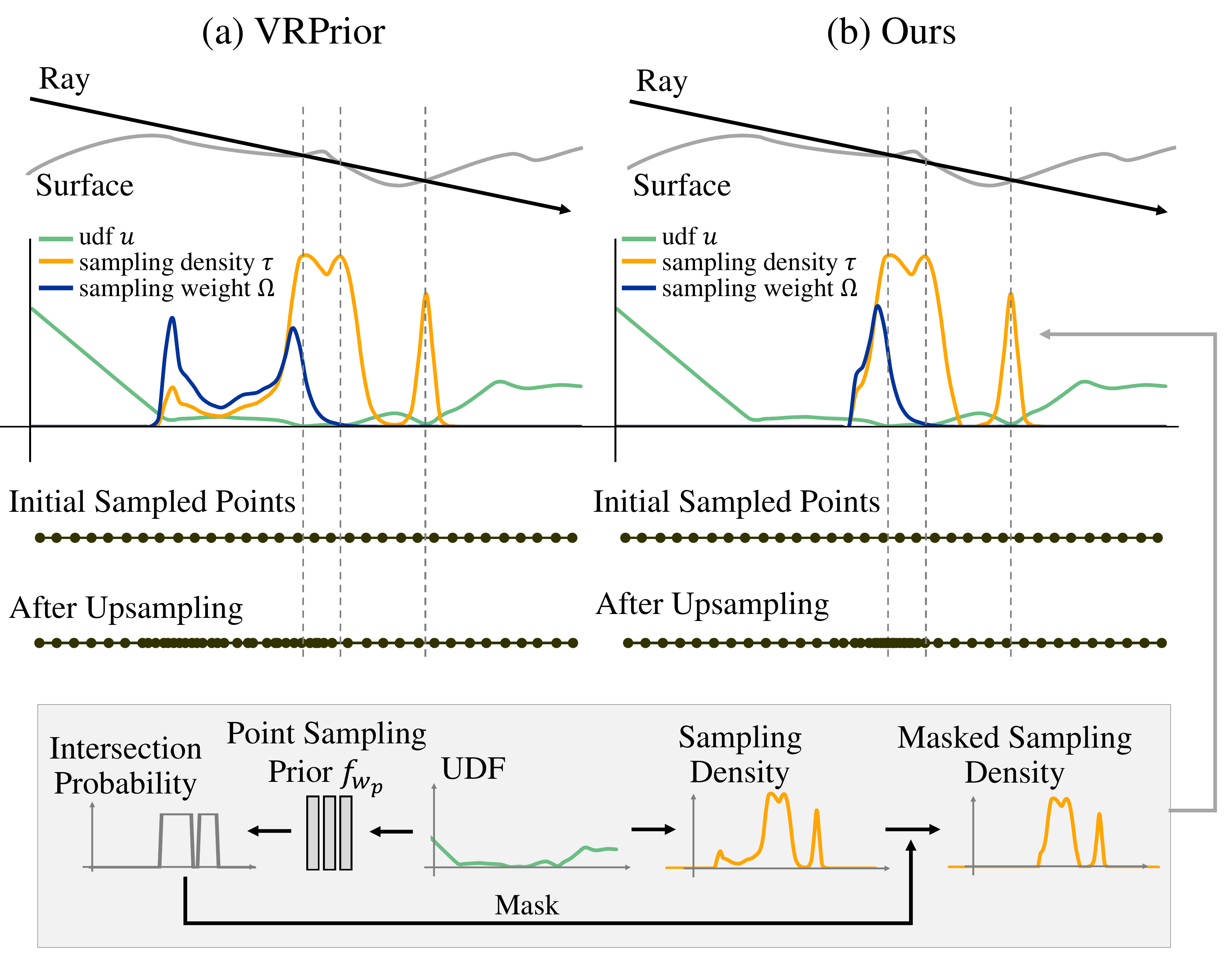}
  \vspace{-0.2cm}
  \caption{Illustration of our UDF upsampling strategy. We significantly mitigate the bias in hierarchical sampling by introducing an auxiliary point sampling prior to correct the sampling density in non-intersection regions.}
  \label{fig:intersectionprior}
\end{figure}
\begin{figure}[t]
  \centering
  \includegraphics[width=\linewidth]{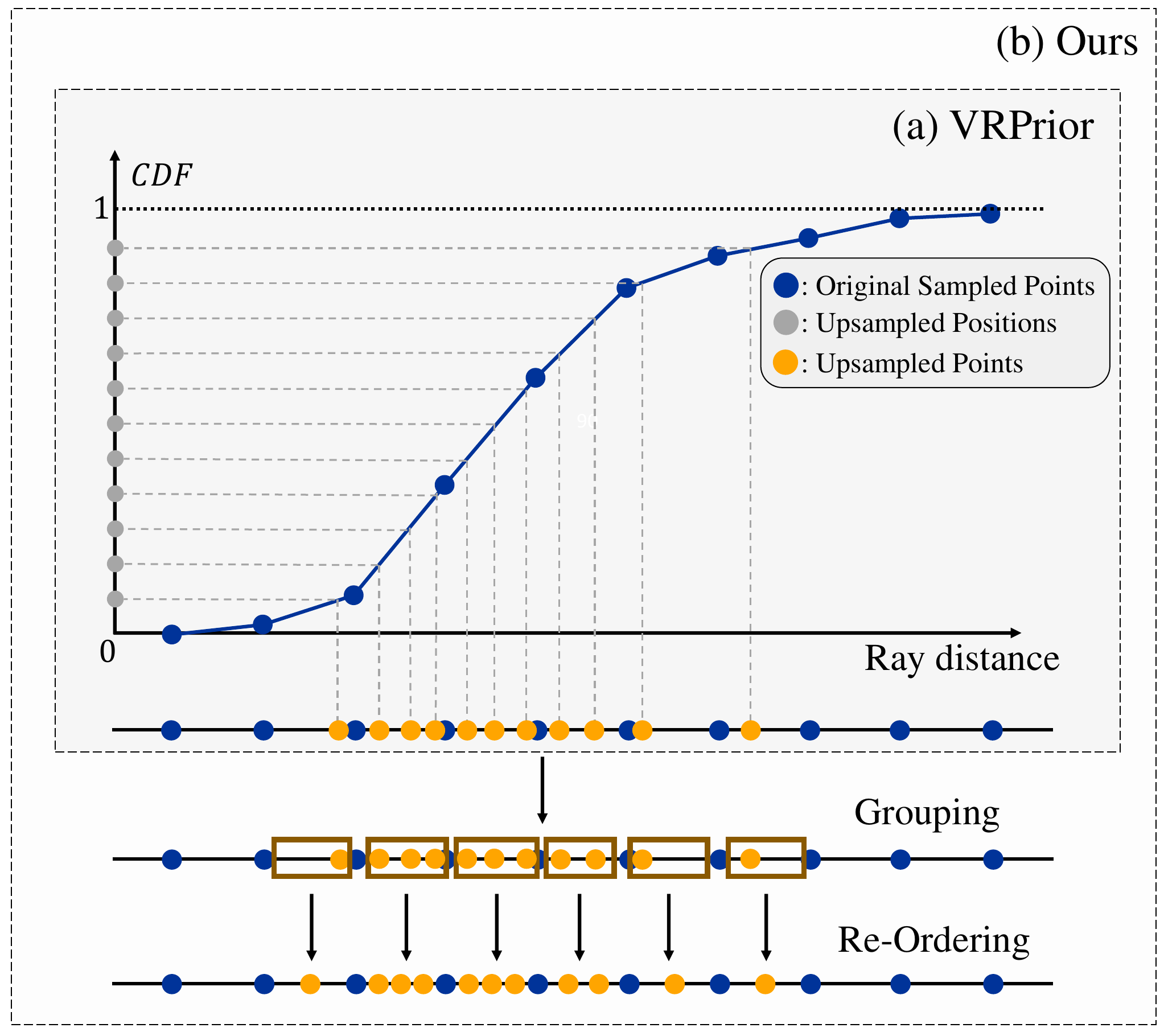}
  \caption{Illustration of our uniform upsampling strategy. We group the upsampled points that fall within the same interval formed by the original sampled points, and re-order them to evenly distribute them within the interval.}
  \label{fig:cdf-uniform-sampling}
\end{figure}

\subsection{Hierarchical Volume Sampling for UDFs}
\label{sec3.5}
Hierarchical volume sampling is an important strategy of improving rendering efficiency and 3D perception in volume rendering. NeuS~\cite{wang2021neus} uses S-density function with fixed standard deviations for importance sampling, while the situation becomes more complex when it comes to UDF, as UDF which is unlike SDF, lacks an explicit distinction of zero-level set. Recent solutions~\cite{liu2023neudf, zhang2024learning} propose a tailored weight function for UDF hierarchical sampling. Specifically, to upsample $N_2$ new points based on the original sampled points $\{q_n\}_{n=1}^{N_1}$, the positions of the upsampled points obey the posterior probability distribution,
\begin{equation}
\begin{split}
\Omega(n)&=(1-e^{-\tau_n\delta_n}) \times \prod_{n'=1}^{n-1}e^{-\tau_{n'}\delta_{n'}}, \\
\tau_n&=\frac{s\cdot e^{-s\cdot f_u(q_n)}}{(1+e^{-s\cdot f_u(q_n)})^2},
\end{split}
\end{equation}
where the sampling weight $\Omega_n$ is the cumulative probability over the interval $[q_n, q_{n+1}]$, that is, the probability that a newly generated upsampled point falls between $q_n$ and $q_{n+1}$. $\delta_n$ is the interval between $q_n$ and $q_{n+1}$, $\tau_n$ denotes the sampling density at $q_n$ and $s$ is a fixed standard deviation, respectively. We observe that this design introduces significant bias before the ray-surface intersection, which is primarily due to the lack of 3D awareness, similar as the analysis in Sec.~\ref{sec3.2}. As the example shown in Fig.~\ref{fig:intersectionprior} (a), due to the cumulative multiplicative effect, the high sampling densities before the ray-surface intersection result in significantly diminished sampling weights at the actual intersection. The bias of sampling weights consequently induces a concentrated distribution of upsampled points in the erroneous region before the ray-surface intersection.

\begin{figure}[t]
  \centering
  \includegraphics[width=\linewidth]{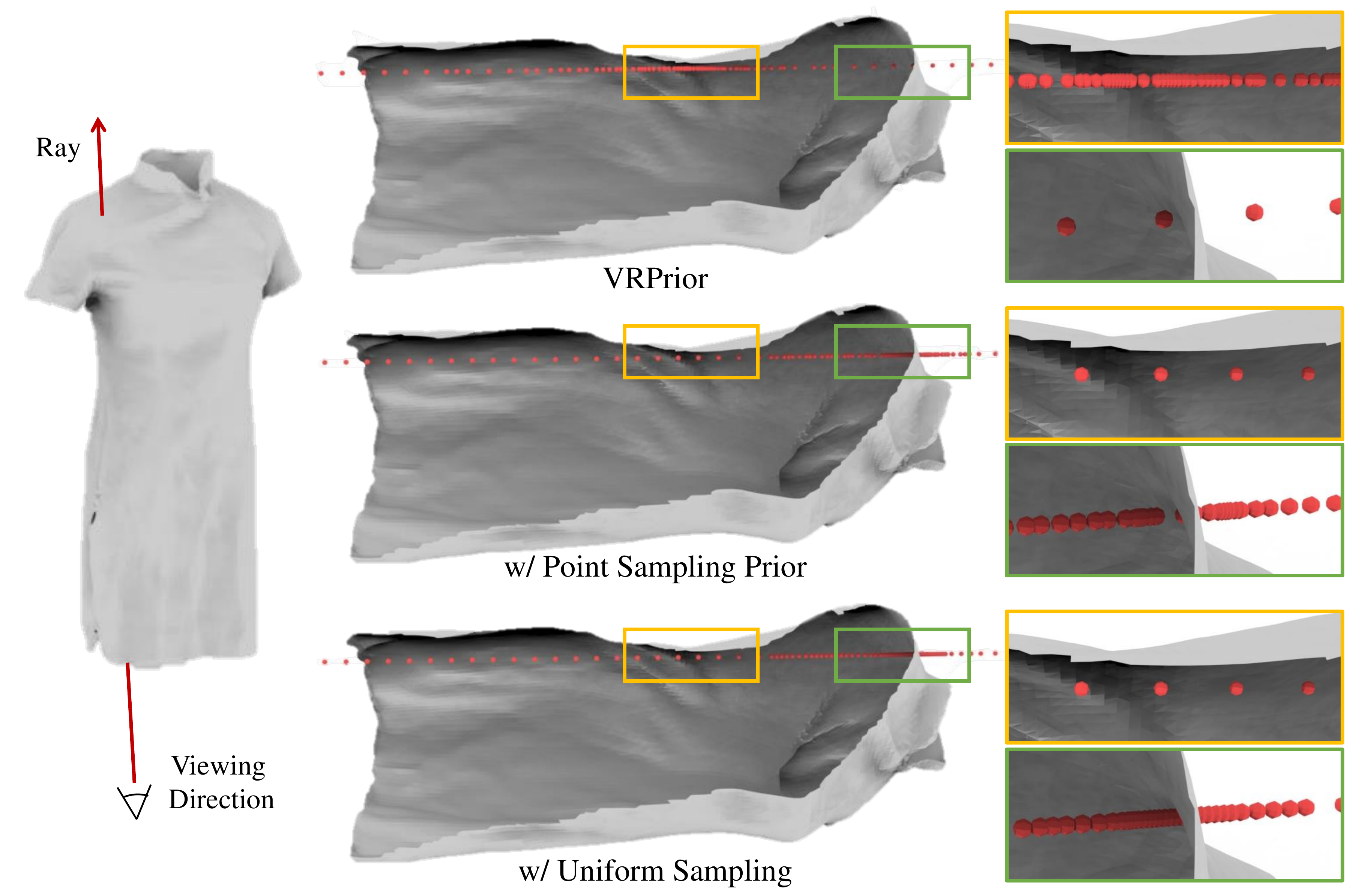}
  \caption{Visualization of upsampled points based on our proposed point sampling prior and uniform upsampling strategy.}
  \label{fig:upsampling-visual}
  \vspace{-0.1cm}
\end{figure}

To mitigate the bias in hierarchical sampling, we introduce an auxiliary point sampling prior $f_{w_p}$ as a mask function of sampling densities. Specifically, the point sampling prior takes a window of UDFs as input and outputs the probability that the window contains a ray-surface intersection. The training of $f_{w_p}$ is similar as that of $f_w$, while the difference is that we use mask images as optimizing objectives here because we want $f_{w_p}$ to produce binary outputs. The mask value is then multiplied with $\{\tau_n\}$ for subsequent hierarchical sampling. The reason that we do not directly use the trained volume rendering prior for $\{\tau_n\}$ is that the training process of volume rendering prior itself also relies on hierarchical sampling. As demonstrated in Fig.~\ref{fig:intersectionprior} (b), the refined sampling densities $\{\tau_n\}$ through our point sampling prior $f_{w_p}$ effectively concentrates the upsampled points near the intersection.

Furthermore, we delve into the inverse transform sampling process and introduce a regularization term to promote the uniform distribution of upsampled points. Given uniformly distributed values $\{u_n\}_{n=1}^{N_2}$ on $[0,1]$, the positions of the $N_2$ upsampled points are determined by $F^{-1}_\Omega(u_n)$, where $F^{-1}_\Omega$ is the inverse cumulative distribution function (CDF) of the sampling weights $\Omega$. Since $F^{-1}_\Omega$ is a non-smooth step function under the situation of discrete sampling, the distribution of the upsampled points $\{q'_n\}_{n=1}^{N_2}$ along the ray is highly uneven, which results in a perception degeneration of the radiance fields, as shown in Fig.~\ref{fig:cdf-uniform-sampling} (a). To facilitate the uniformity of the upsampling process, we first group the newly generated points that fall within the same interval of the original sampled point as $\{q'_{n,1},q'_{n,2},...q'_{n,m}\}$, where $q'_{n,j}$ represents the $j$-th point that fall within $[q_n, q_{n+1}]$. Then we re-order the $m$ points such that the interval of consecutive points $q'_{n,j}$ and $q'_{n,j+1}$ is $\frac{\delta_n}{m+1}$, where $\delta_n$ is the distance between $q_n$ and $q_{n+1}$, and $m$ is the number of points that fall within $[q_n, q_{n+1}]$, as illustrated in Fig.~\ref{fig:cdf-uniform-sampling} (b). Through this way, we ensure the uniform distribution of upsampled points, thereby facilitating a comprehensive perception in 3D space during sampling. 

We provide an example of sampling on a real garment in Fig.~\ref{fig:upsampling-visual}, where the ray starts from the bottom of the garment, passes through the waist, and eventually exits at the shoulder. Sampling bias causes VRPrior to erroneously concentrate the upsampled points around the waist of the garment. Guided by our point sampling prior, the upsampling process successfully circumvents the pseudo ray-surface intersection. Eventually, with our uniform sampling strategy, the upsampled points are densely and uniformly distributed near the ray-surface intersection.

\begin{figure}[t]
  \centering
  \includegraphics[width=\linewidth]{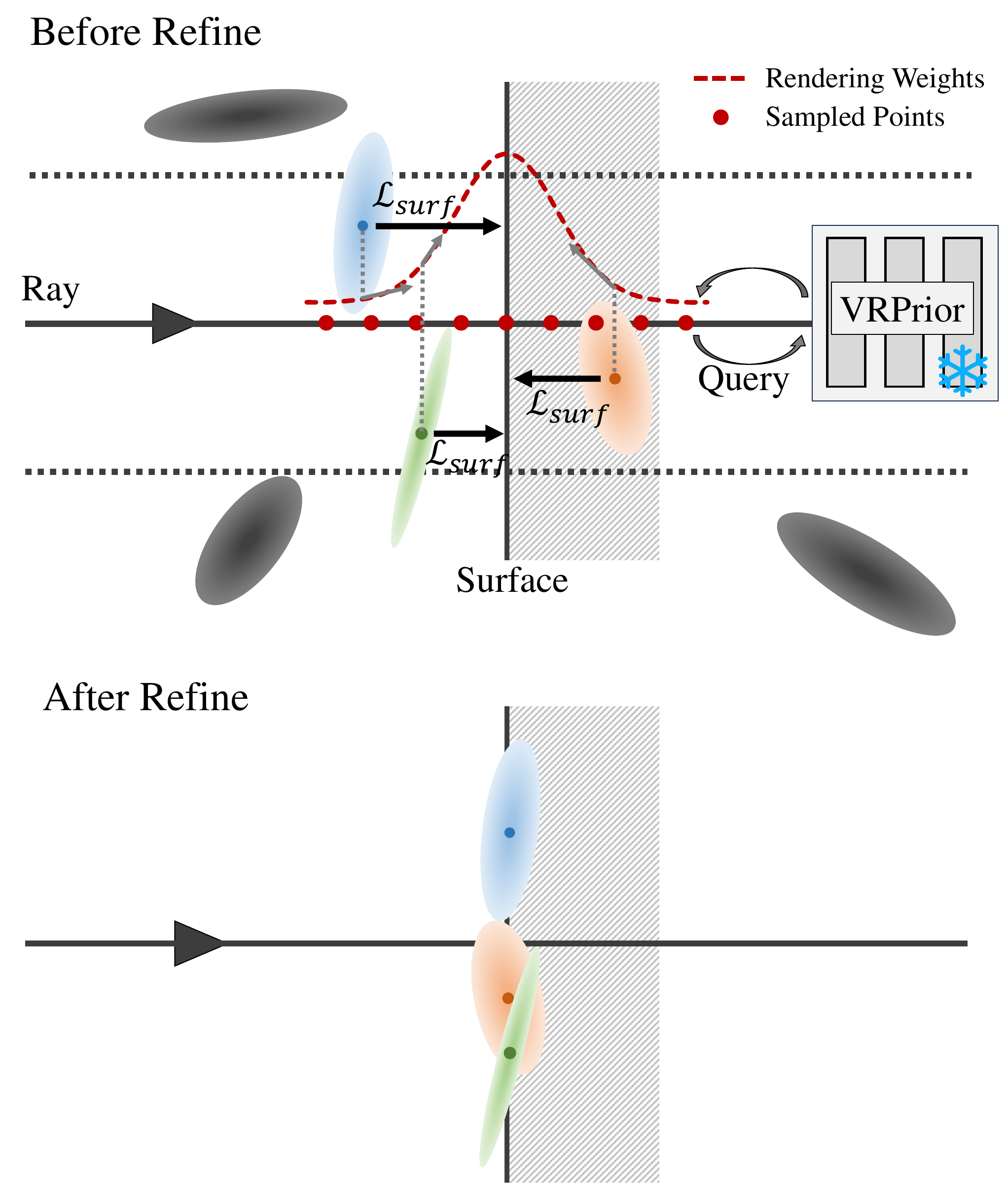}
  \vspace{-0.1cm}
  \caption{Visualization of our strategy of refining Gaussian Splatting using pretrained volume rendering priors. The Gaussians are pushed toward the direction of increasing rendering weights under dense queries.}
  \label{fig:vrprior+gs}
\end{figure}

\subsection{Refining Gaussian Splatting with VRPrior}
\label{sec3.6}
Recently, Gaussian Splatting has achieved remarkable progress in novel view synthesis~\cite{kerbl20233dgs,yu2024mipsplatting,han2024binocular}, yet Gaussian-based surface reconstruction still struggles with discreteness, sparseness, and off-surface drift. Although several methods alleviate these issues by introducing regularization terms to align Gaussians with the surface~\cite{zhang2024gspull,li2025gaussianudf,huang20242dgs,chen2024pgsr}, they either cannot be integrated with neural implicit representations or incur the degeneration of geometric details. The reason is that these constraints are self-distilled from the Gaussians themselves and lack explicit knowledge of the guidance toward true surfaces. To address this limitation, we propose to employ our pretrained volume rendering priors as a general Gaussian surface prior that pushes the Gaussians toward the true surfaces predicted by the volume rendering prior, thereby refining the reconstructed geometry.

Specifically, as illustrated in Fig.~\ref{fig:vrprior+gs}, given a trained Gaussian Splatting model, we optimize an unsigned distance function on it, like GaussianUDF~\cite{li2025gaussianudf}. The UDF field provides a coarse estimation of the surface from the discrete and sparse Gaussian representations. We emit rays $V_k$ from the training views $\{I_j\}_{j=1}^J$, uniformly sample query points $\{q_n\}$ along each ray to obtain their UDF values and compute the rendering weights distribution $\{w_n\}$ around the surface. We do not use hierarchical sampling here in order to avoid fluctuations caused by random sampling intervals. We then select Gaussians $\{G_t\}$ that are both close to the ray and near the surface, interpolate the rendering weights at their projected positions on the ray, and compute the direction in which the rendering weight increases. A surface loss is applied to the Gaussians to move them along the direction of increasing rendering weights,
\begin{equation}
\begin{gathered}
    \mathcal{L}_{surf}=\Vert \mu_t - \text{sg}(\mu_t+r_k*(d^*-d_t))\Vert , \\
    d^*=d_{n^*}, \ n^*=\arg\max_n w_n
\end{gathered}
\end{equation}
\noindent where $\mu_t$ is the center of Gaussian $G_t$, $r_k$ is the direction of ray $V_k$, $d_t$ is the depth of the projection point from $G_k$ onto $V_k$, $d_n$ is the depth of $\{q_n\}$ and $\text{sg}(\cdot)$ is stop-gradient operation. Although the UDFs are not perfectly accurate, they provide reliable optimization directions for refining the Gaussian distribution. The Gaussian model and UDF model are jointly finetuned in a progressive manner, continuously improving each other and refining the geometry over time. Compared with existing methods that directly pull Gaussians to the UDF zero-level set~\cite{li2025gaussianudf,zhang2024gspull}, our approach leverages dense queries around the zero-level set as strong geometric cues. It refines the surface in a conservative and gradual manner, effectively preventing error accumulation caused by ambiguous pulling directions.

%-----------------------------------------------------
%-----------------------------------------------------
\section{Experiments}

\subsection{Experiment Settings}
\subsubsection{Data for Learning Priors}
We select one object from each one of ``car'', ``airplane'', ``bottle'' categories in ShapeNet dataset~\cite{chang2015shapenet} and three objects from DeepFashion3D dataset~\cite{zhu2020deepfashion3d} to form our training dataset for learning volume rendering priors. There is no overlap between the selected object and the testing objects. Our ablation studies demonstrate that these objects are sufficient to learn accurate volume rendering priors with good generalization capabilities across various shape categories. For each object, we first convert it into a normalized watertight mesh and then render 100 depth images with 600$\times$600 resolution from uniformly distributed viewpoints on a unit sphere. We utilize the volume rendering priors pre-trained on these six shapes to report our results.

\subsubsection{Datasets for Evaluations}
We evaluate our method on five datasets including DeepFashion3D (DF3D)~\cite{zhu2020deepfashion3d},  DTU~\cite{jensen2014large}, Replica~\cite{straub2019replica}, Insects and real-captured datasets. For DF3D dataset, we follow~\cite{long2023neuraludf} and use the same 12 garments from different categories. For DTU dataset, we use the same 15 scenes that are widely used by previous studies. And we use all the 8 scenes in Replica dataset. To further evaluate the effectiveness of our method on reconstructing open structures, we select 10 insect objects from the Objaverse-XL dataset~\cite{deitke2023objaverse-xl} to form the Insects dataset~\cite{zhou2025udfstudio}. These objects feature detailed geometric elements as well as thin structures and open surfaces such as wings, which presents a challenge for multi-view reconstruction with UDFs. Similar to the process of constructing data for learning priors, we render 100 RGB images with 1024$\times$1024 resolution from surrounding viewpoints for each object to form the training set. We also report results on real scans from NeUDF~\cite{liu2023neudf} and the ones shot by ourselves.

\subsubsection{Baselines}
We compare our method with the state-of-the-art methods which use different differentiable renderers to reconstruct open surfaces, including NeuralUDF~\cite{long2023neuraludf}, NeUDF~\cite{liu2023neudf}, NeAT~\cite{meng2023neat}, 2S-UDF~\cite{deng20242sudf} and our previous work VRPrior~\cite{zhang2024learning}. We also compare our methods with state-of-the-art Gaussian Splatting-based reconstruction methods, including 2DGS~\cite{huang20242dgs}, GOF~\cite{yu2024gof}, GaussianUDF~\cite{li2025gaussianudf}, PGSR~\cite{chen2024pgsr}, GeoSVR~\cite{li2025geosvr} and QGS~\cite{zhang2024quadratic}. 

\subsubsection{Metrics}
For DTU dataset and DF3D dataset, we use Chamfer Distance (CD) as the metric. For Replica dataset, we report CD, Normal Consistency (N.C.) and F1-score following previous works~\cite{yu2022monosdf, azinovic2022neuralrgbd}. Moreover, we report the rendering errors in Tab.~\ref{tab:depth-error-metric} using depth L1 distance, mask errors with cross entropy and L1 distance.

\subsubsection{Implementation Details}
Our volume rendering prior network consists of MLPs with 256 hidden units, where $f_{w_1}$ comprises 3 layers and $f_{w_2}$ has 6 layers with skip connections. The UDF function $f_u$ is an 8-layer MLP with skip connections and the color function $f_c$ is a 2-layer MLP with 256 hidden units. We apply Softplus as the activation function after the last layer of $f_u$ to preserve the output non-negative. The learning rate and decay scheduler of $f_u$ and $f_c$ are the same as previous works~\cite{long2023neuraludf, wang2021neus}. We remove the normal vectors from the input of the color network because the normals are ambiguous at the zero-level set due to the non-differentiability of UDFs. \textcolor{red}{For the experiments described in Sec.~\ref{sec3.6}, we train the Gaussian Splatting methods for 30k iterations. We begin training the UDFs from 9k iteration. During the final 10k iterations, we refine Gaussians by casting rays with a batch size of 2048, where 64 points are sampled along each ray.}

\begin{table}[t]
  \centering
  \caption{Numerical comparisons in all ShapeNet categories. ``Train'' denotes the results evaluated on VRP training sets.}
  \vspace{-0.2cm}
  \label{tab:depth-error-metric}
  \resizebox{\linewidth}{!}{\begin{tabular}{l|cccc}
    \toprule
     Methods                            & Depth-L1$\downarrow$ & Mask-Entropy$\downarrow$ & Mask-L1$\downarrow$ & Peak Diff-L1$\downarrow$  \\
    \midrule
     NeuS-UDF\cite{wang2021neus}       & 3.46$\pm$1.49 & 2.67$\pm$1.27 & 2.27$\pm$1.09 & 17.02$\pm$2.53     \\
     NeUDF\cite{liu2023neudf}          & 1.67$\pm$0.31 & 4.19$\pm$2.02 & 0.89$\pm$0.14 & 2.03$\pm$0.35    \\
     NeuralUDF\cite{long2023neuraludf} & 1.28$\pm$0.28 & 30.65$\pm$7.59 & 0.61$\pm$0.12 & 2.36$\pm$0.51     \\
     NeuS-SDF\cite{wang2021neus}       & 0.97$\pm$0.67 & 0.60$\pm$0.56 & 0.52$\pm$0.47 & 2.07$\pm$0.29 \\
     2S-UDF\cite{deng20242sudf}        & 0.45$\pm$0.21 & 2.31$\pm$1.05 & 0.17$\pm$0.08 & 1.48$\pm$0.46    \\
     VRPrior\cite{zhang2024learning}   & 0.41$\pm$0.19 & 0.70$\pm$0.63 & 0.12$\pm$0.05 & 2.34$\pm$0.58    \\
     Ours                              & \textbf{0.33$\pm$0.14} & \textbf{0.57$\pm$0.55} & \textbf{0.09$\pm$0.04} & \textbf{1.32$\pm$0.37}    \\
   \cmidrule{1-5} 
     NeuS (Train)                      & 0.88$\pm$0.71 & 0.54$\pm$0.49 & 0.59$\pm$0.40 & 2.09$\pm$0.21    \\
     Ours (Train)                      & \textbf{0.11$\pm$0.03} & \textbf{0.31$\pm$0.28} & \textbf{0.05$\pm$0.04} & \textbf{0.95$\pm$0.22}    \\
  \bottomrule
    \end{tabular}}
\end{table}
\begin{table}[t]
  \centering
  \caption{Quantitative evaluations on DF3D\cite{zhu2020deepfashion3d}, DTU\cite{jensen2014large} and Replica\cite{straub2019replica} datasets. }
  \label{tab:df3d-dtu-replica-compare}
  \resizebox{\linewidth}{!}{\begin{tabular}{l|c|c|ccc}
    \toprule
    Datasets & DF3D & DTU & \multicolumn{3}{c}{Replica} \\
    \midrule
     Metrics & CD$\downarrow$ & CD$\downarrow$ & CD$\downarrow$ & N.C.$\uparrow$ & F-score$\uparrow$  \\
    \cmidrule{1-6}
     COLMAP\cite{schonberger2016structure} & 3.10 & 1.36 & 0.23 & 0.46 & 0.43     \\
     NeuS\cite{wang2021neus}               & 4.36 & 0.87 & 0.07 & 0.88 & 0.69      \\
     NeuralUDF\cite{long2023neuraludf}     & 2.15 & 0.75 & 0.11 & 0.85 & 0.53      \\
     NeAT\cite{meng2023neat}               & 2.10 & 0.88 & 0.18 & 0.75 & 0.36     \\
     NeUDF\cite{liu2023neudf}              & 2.01 & 1.58 & 0.28 & 0.78 & 0.31      \\
     2S-UDF\cite{deng20242sudf}            & 1.99 & 0.90 & 0.34 & 0.63 & 0.16      \\
     VRPrior\cite{zhang2024learning}       & 1.71 & 0.85 & 0.04 & 0.90 & 0.80     \\
   \cmidrule{1-6}
     2DGS\cite{huang20242dgs}              & 3.81 & 0.80 & 0.13 & 0.85 & 0.57 \\
     GOF\cite{yu2024gof}                   & 2.49 & 0.74 & 0.09 & 0.80 & 0.58 \\
     GaussianUDF\cite{li2025gaussianudf}   & 1.60 & 0.68 & 0.07 & 0.91 & 0.77 \\ 
     \textcolor{red}{PGSR\cite{chen2024pgsr}} & \textcolor{red}{3.25} & \textcolor{red}{0.56} & \textcolor{red}{0.07} & \textcolor{red}{0.90} & \textcolor{red}{0.83} \\
     \textcolor{red}{GeoSVR\cite{li2025geosvr}} & \textcolor{red}{2.91} & \textcolor{red}{\textbf{0.47}} & \textcolor{red}{0.08} & \textcolor{red}{0.90} & \textcolor{red}{0.82} \\
     QGS\cite{zhang2024quadratic}          & 4.46 & 0.54 & 0.06 & 0.91 & 0.74   \\
   \cmidrule{1-6} 
     Ours                                  & \textbf{1.59} & 0.66 & \textbf{0.03} & \textbf{0.92} & \textbf{0.85}     \\
  \bottomrule
    \end{tabular}}
\end{table}
\begin{figure*}[t]
  \centering
  \includegraphics[width=\linewidth]{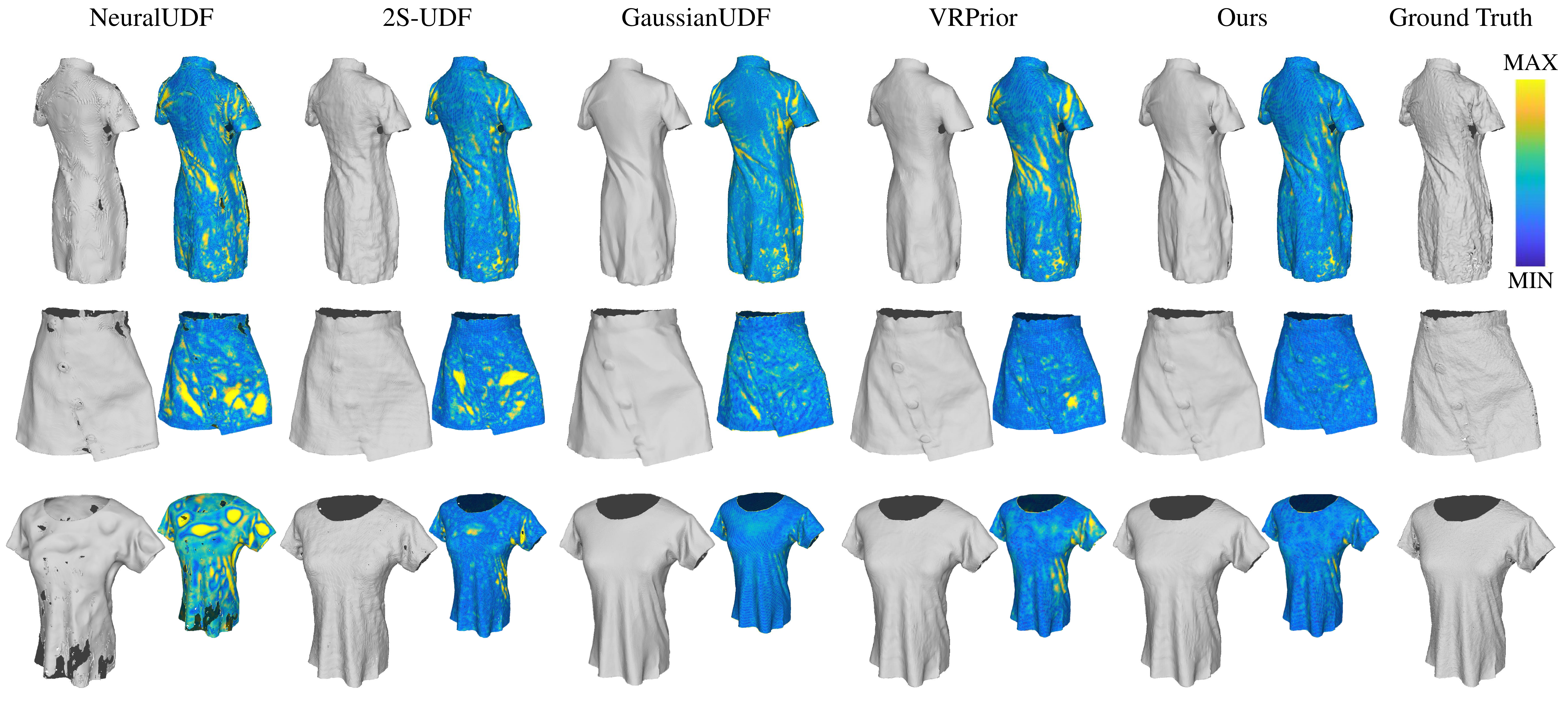}
  \vspace{-0.3cm}
  \caption{Visual comparisons on open surface reconstructions with error maps on DeepFashion3D~\cite{zhu2020deepfashion3d} dataset.}
  \label{fig:df3d-visual-compare}
\end{figure*}
\begin{figure*}[t]
  \centering
  \includegraphics[width=\linewidth]{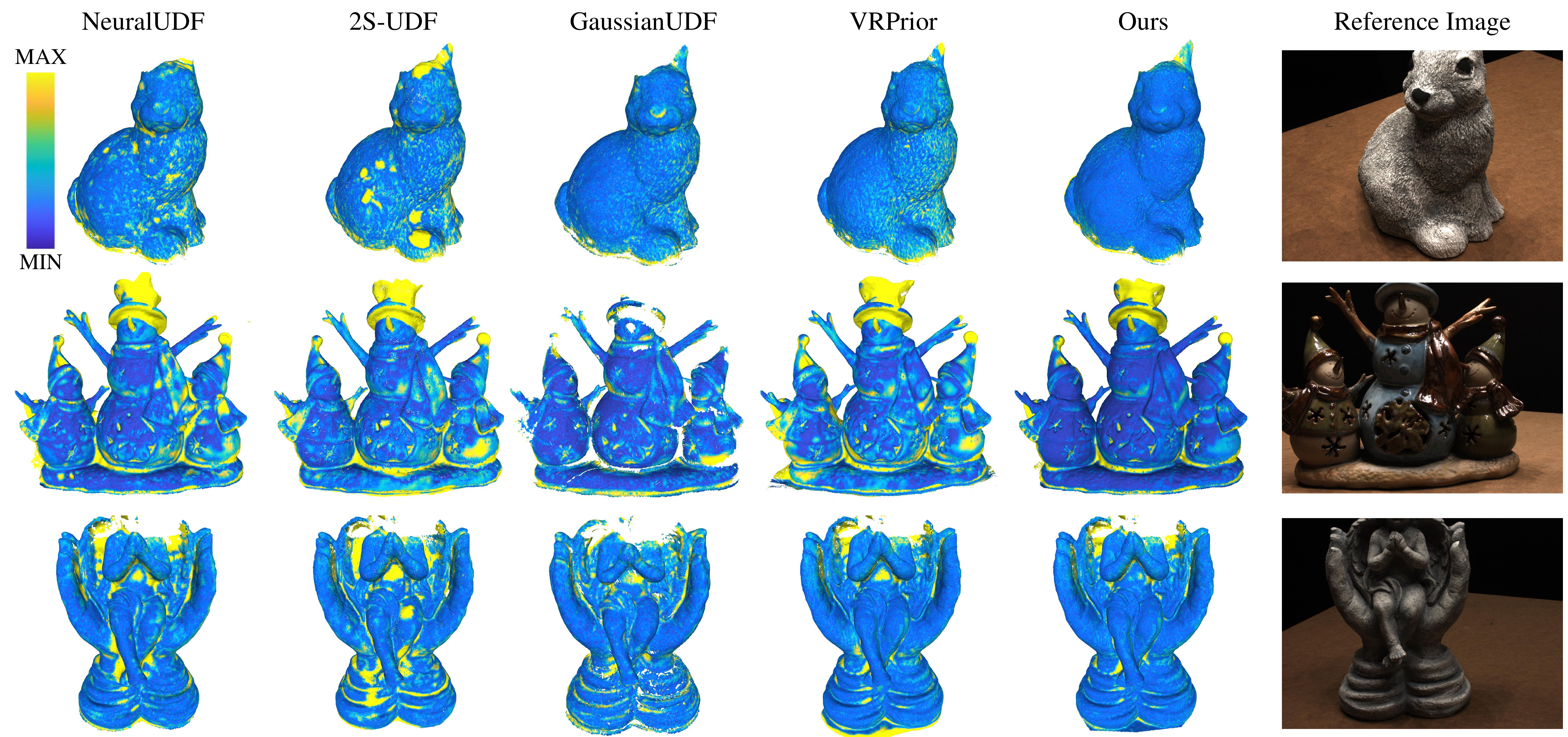}
  \vspace{-0.3cm}
  \caption{Visual comparisons of error maps on DTU~\cite{jensen2014large} dataset.}
  \label{fig:dtu-visual-compare}
\end{figure*}
\vspace{-0.2cm}

\begin{figure*}[t]
  \centering
  \includegraphics[width=\linewidth]{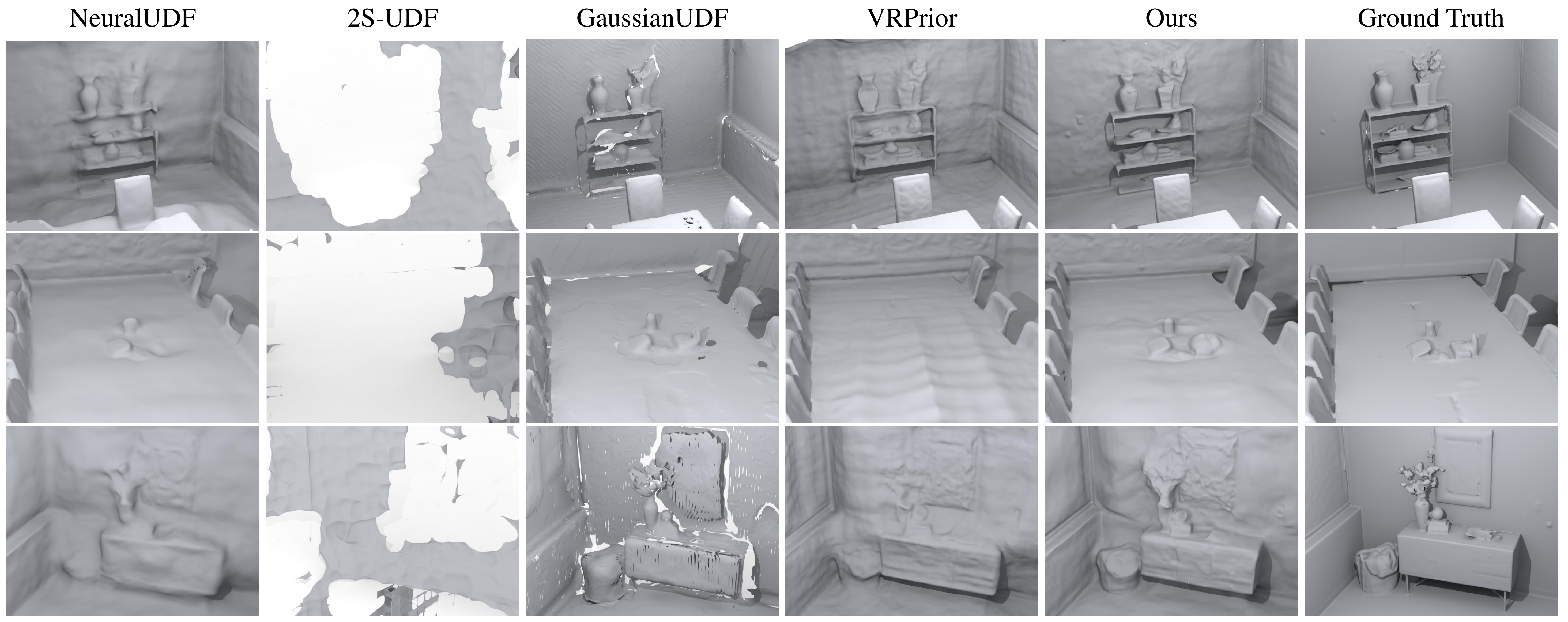}
  \vspace{-0.1cm}
  \caption{Qualitative comparisons on Replica~\cite{straub2019replica} dataset. Our method outperforms other methods on complex indoor scenes while other UDF-based methods struggle to recover complete and smooth surfaces.}
  \label{fig:replica-visual-compare}
\end{figure*}
\begin{figure*}[t]
  \centering
  \includegraphics[width=\linewidth]{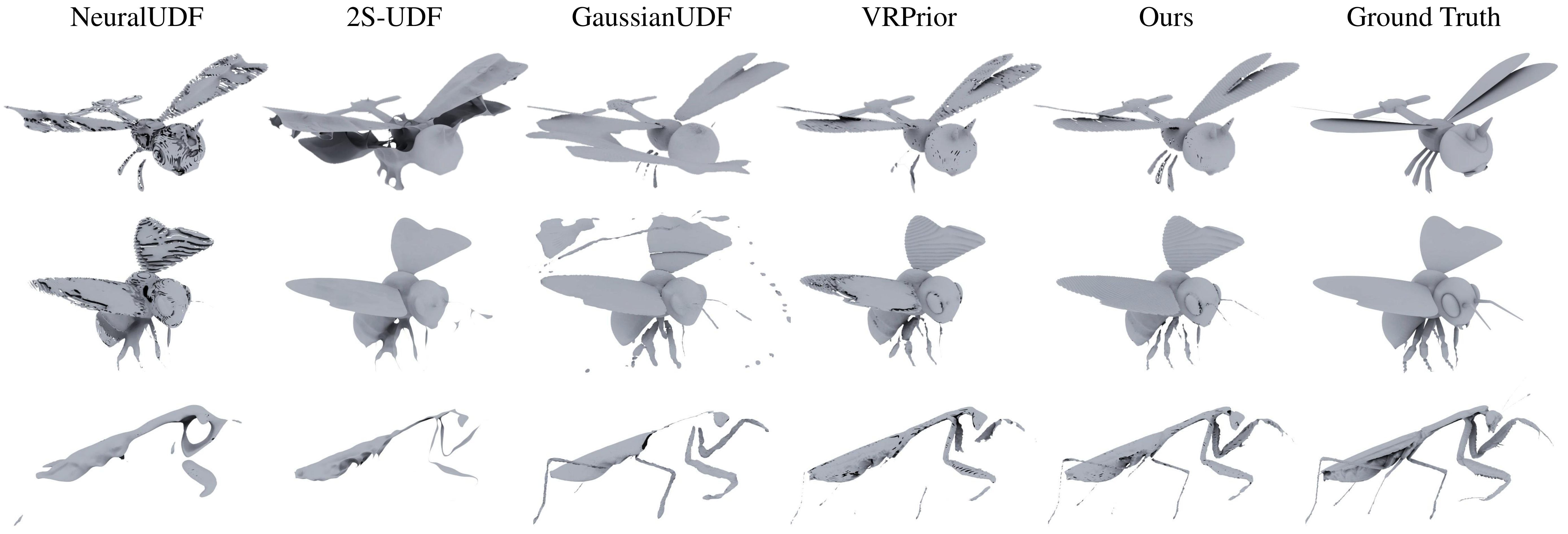}
  \vspace{-0.1cm}
  \caption{Visual comparisons of reconstructed meshes on Insects dataset. Our method shows clear advantages in reconstructing thin structures such as the wings and the legs.}
  \label{fig:insects-visual-compare}
\end{figure*}
\begin{figure}[t]
  \centering
  \includegraphics[width=\linewidth]{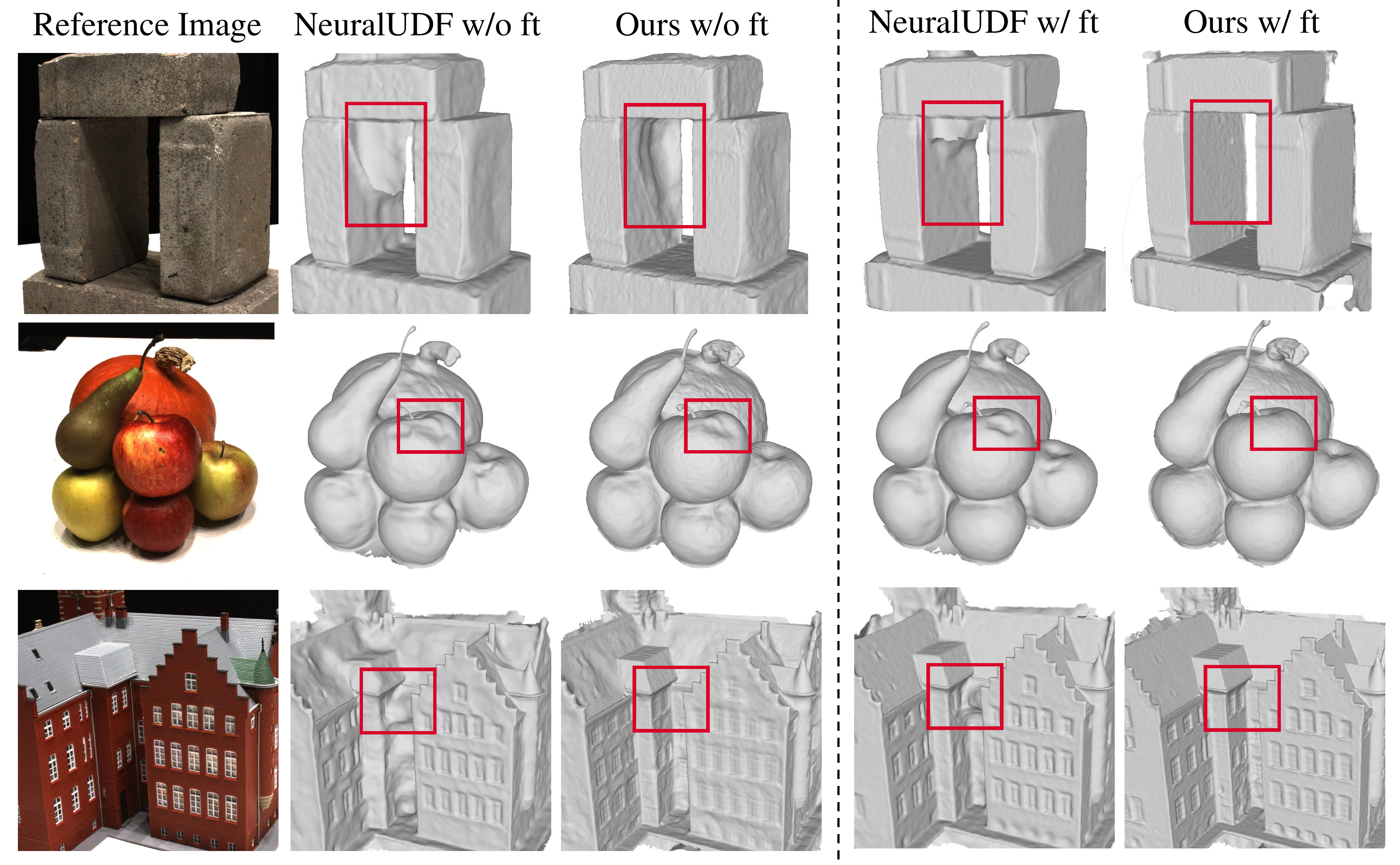}
  \caption{Visual Comparisons between our method and NeuralUDF~\cite{long2023neuraludf} in DTU~\cite{aanaes2016large} dataset. Our method, when fine-tuned using patch loss, still outperforms NeuralUDF under the same experimental conditions.}
  \label{fig:dtu-patchloss-compare}
\end{figure}
\begin{table}[t]
  \centering
  \caption{Numerical results on DTU with patch loss.}
  \label{tab:dtu-patchloss}
  \resizebox{1.0\linewidth}{!}{
  \begin{tabular}{l|ccc}
    \toprule
    Methods & Accuracy$\downarrow$ & Completion$\downarrow$ & Mean$\downarrow$  \\
    \midrule
     NeuralUDF\cite{long2023neuraludf}  & 1.05 & 1.10 & 1.07      \\  
     Ours                           & \textbf{0.72} & \textbf{0.89} & \textbf{0.80} \\
    \cmidrule{1-4}
     NeuralUDF (w/ ft)              & 0.83 & 0.67 & 0.75      \\
     Ours (w/ ft)                   & \textbf{0.73} & \textbf{0.59} & \textbf{0.66} \\
  \bottomrule
    \end{tabular}}
\end{table}

\subsection{Comparisons with the Latest Methods}

\subsubsection{Results on ShapeNet} Tab.~\ref{tab:depth-error-metric} reports numerical comparisons in the experiment in Fig.~\ref{fig:depth-error-curve} in terms of 3 metrics. We report the averages and variances over all 55 shapes. All of the objects are normalized to the range [-1, 1], and the values are magnified 100 times. We render depth images and mask images by forwarding ground truth unsigned distances at the same set of queries as other methods with the learned prior. For the learnable convergence parameters in the baseline methods, we fix them to the average values observed at convergence on DTU dataset. These values provide a reasonable reference for the choice of the factors. It should be noted that a larger factor does not correspond to higher accuracy. For instance, the factor in NeuralUDF determines the location where the UDF is flipped into an SDF. A very large value may cause this flip to occur before true surface, resulting in depth rendering errors. We calculate the L1-error and cross entropy error between predicted images and ground truth ones. We also measure the depth discrepancy between the sampled point with maximum rendering weight and the true surface along each ray, denoted as ``Peak Diff-L1''. We achieve the best accuracy among all renderers for UDFs and SDFs. Although our volume rendering prior is unbiased since it is trained with ground truth UDFs and depths, certain inaccuracies still occur in real rendering due to random hierarchical sampling and the ill definition of the UDF field near the surfaces. This phenomenon is common in all volume rendering frameworks. Even NeuS~\cite{wang2021neus}, which employs a theoretically unbiased mapping, exhibits noticeable depth inaccuracies in practical scenarios, as shown in the ``Peak Diff-L1'' column of Tab.~\ref{tab:depth-error-metric}. We also report the results of ``NeuS-SDF'' and our method on VRP training sets, which demonstrate the strong generalization ability of our approach.

\begin{figure}[t]
  \centering
  \includegraphics[width=\linewidth]{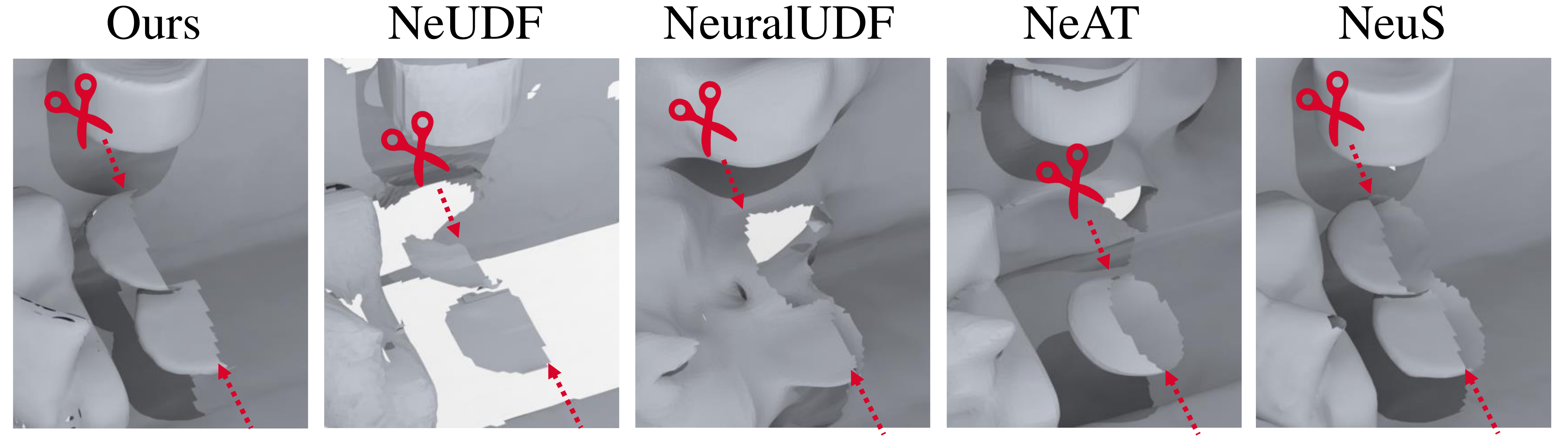}
  \caption{Illustration of the capabilities of reconstructing single-layer geometries in indoor scenes.}
  \label{fig:ablation-single-face}
\end{figure}

\subsubsection{Results on DF3D} We first report evaluations on DF3D (CD $\times 10^{-3}$). Numerical comparison in Tab.~\ref{tab:df3d-dtu-replica-compare} indicates that our learned prior produces the lowest CD errors among all handcrafted renderers. The visual comparison in Fig.~\ref{fig:df3d-visual-compare} details our superiority on reconstruction with error maps. We see that our prior helps the network to recover not only smoother surfaces at most areas but also sharper edges at wrinkles.

\begin{table*}[t]
  \centering
  \caption{Reconstruction Comparison in terms of Chamfer Distance ($\times 10^{-3}$) on Insects datasets.}
  \label{tab:insects-compare}
  \resizebox{.95\linewidth}{!}{\begin{tabular}{l|cccccccccc|c}
    \toprule
    Methods & S00 & S01 & S02 & S03 & S04 & S05 & S06 & S07 & S08 & S09 & Mean \\
    \midrule
     NeuS\cite{wang2021neus}              & 3.22 & 4.61 & 3.63 & 5.93 & 5.94 & 4.75 & \textbf{2.84} & 3.79 & 4.32 & 6.90 & 4.59  \\
     NeuralUDF\cite{long2023neuraludf}    & 3.05 & 4.65 & 10.16 & 10.16 & 9.56 & 3.48 & 3.68 & 2.77 & 3.13 & \textbf{1.52} & 5.22 \\
     NeAT\cite{meng2023neat}              & 6.40 & 5.74 & 2.85 & 12.15 & 6.32 & 4.94 & 3.98 & 5.71 & 7.21 & 6.31 & 6.16   \\
     NeUDF\cite{liu2023neudf}             & 6.24 & 4.86 & 9.81 & 8.69 & 5.69 & 8.12 & 12.81 & 6.23 & 4.49 & 7.04 & 7.40    \\
     2S-UDF\cite{deng20242sudf}           & 12.16 & 4.83 & 7.15 & 9.49 & 7.40 & 4.89 & 7.71 & 3.20 & 7.98 & 4.21 & 6.90    \\
     VRPrior\cite{zhang2024learning}      & 2.99 & 5.23 & 3.40 & 5.77 & 5.45 & 4.80 & 3.84 & 2.66 & 4.09 & 2.14 & 4.04    \\
    \cmidrule{1-12}
     2DGS\cite{huang20242dgs}             & 3.38 & 3.85 & 5.73 & 3.29 & 3.96 & 2.69 & 1.99 & 4.29 & 2.76 & 6.37 & 3.55  \\
     GOF\cite{yu2024gof}                  & 3.51 & 3.47 & 5.98 & 3.49 & 4.00 & \textbf{2.22} & \textbf{1.74} & 4.48 & \textbf{2.31} & 5.92 & 3.71  \\
     GaussianUDF\cite{li2025gaussianudf}  & 3.64 & 4.57 & 4.31 & 2.88 & 3.87 & 3.19 & 5.18 & 7.13 & 3.77 & 4.89 & 4.34  \\
     \textcolor{red}{PGSR\cite{chen2024pgsr}} & \textcolor{red}{3.25} & \textcolor{red}{4.22} & \textcolor{red}{2.69} & \textcolor{red}{3.32} & \textcolor{red}{3.72} & \textcolor{red}{3.96} & \textcolor{red}{2.78} & \textcolor{red}{3.54} & \textcolor{red}{3.25} & \textcolor{red}{3.99} & \textcolor{red}{3.47}  \\
     \textcolor{red}{GeoSVR\cite{li2025geosvr}} & \textcolor{red}{2.98} & \textcolor{red}{4.01} & \textcolor{red}{2.36} & \textcolor{red}{3.17} & \textcolor{red}{\textbf{3.59}} & \textcolor{red}{3.55} & \textcolor{red}{2.65} & \textcolor{red}{3.63} & \textcolor{red}{3.49} & \textcolor{red}{3.80} & \textcolor{red}{3.32}  \\
     QGS\cite{zhang2024quadratic}         & 3.82 & \textbf{3.40} & 3.17 & \textbf{2.60} & 4.11 & 4.53 & 2.08 & 4.16 & 3.60 & 4.22 & 3.57  \\
   \cmidrule{1-12} 
     Ours                                  & \textbf{2.48} & 4.40 & \textbf{2.31} & 4.47 & 4.26 & 3.71 & 3.34 & \textbf{2.19} & 3.04 & \textbf{1.76} & \textbf{3.19}    \\
  \bottomrule
    \end{tabular}}
\end{table*}
\begin{figure*}
\centering
    \includegraphics[width=\linewidth]{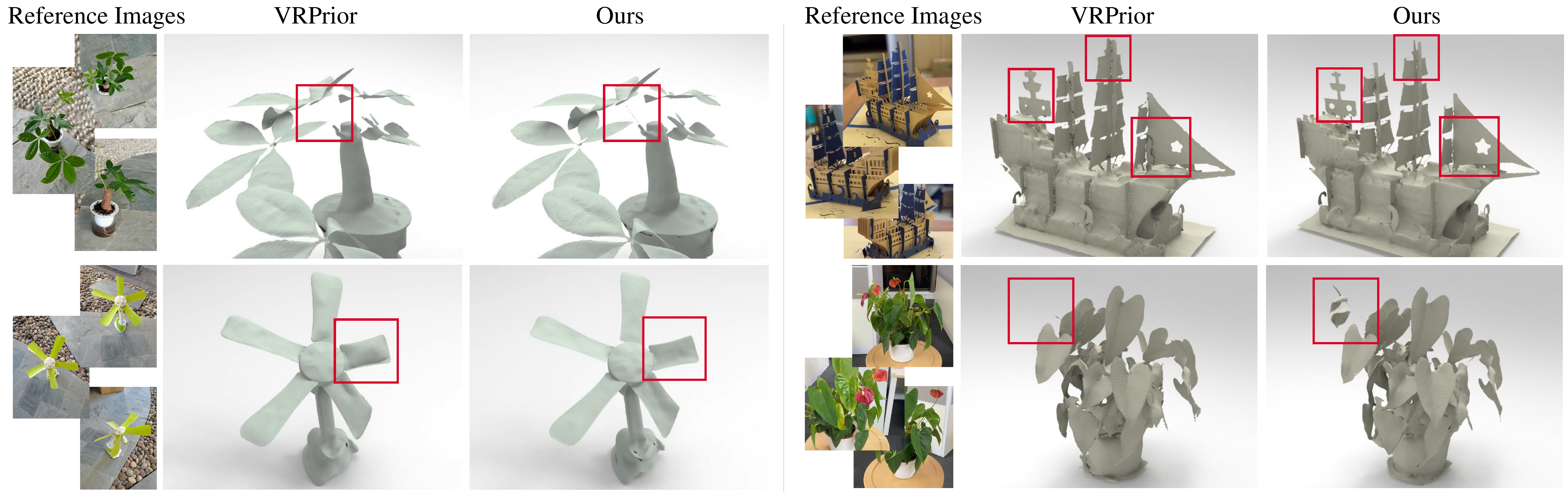}
    \caption{Visual comparisons of real scans between VRPrior and our method.}
  \label{fig:real-plant-fan-visual}
\end{figure*}

\subsubsection{Results on DTU} 
Tab.~\ref{tab:df3d-dtu-replica-compare} reports our evaluation on DTU. Note that NeAT and 2S-UDF rely on additional mask supervision on DTU dataset. Our reconstructions produce low CD errors with our pre-trained prior. Although the shapes used to learn the prior are not related to any scenes in DTU, our prior comes from sets of queries in a local window on a ray, which are more general to unsigned distances along a ray in unobserved scenes. Although some recent 3DGS-based reconstruction methods achieve lower numerical results, they are tailored for closed surfaces and degrades severely in open and thin structures. Visual comparisons in Fig.~\ref{fig:dtu-visual-compare} detail our reconstruction accuracy. Note that NeuralUDF uses additional patch loss~\cite{long2022sparseneus} in DTU dataset to fine-tune the resulted meshes, which is not the primary contribution of the differentiable renderer. Hence, for fair comparison, we report the results of NeuralUDF without fine-tuning across all datasets. However, we also integrate our method with patch loss and report the comparisons in Fig.~\ref{fig:dtu-patchloss-compare} and Tab.~\ref{tab:dtu-patchloss}, including our method and NeuralUDF with or without patch loss fine-tune. The visualization results show that our method, when fine-tuned using patch loss, still outperforms NeuralUDF under the same experimental conditions.

\subsubsection{Results on Replica} We also evaluate our method in indoor scenes. Numerical valuations in Tab.~\ref{tab:df3d-dtu-replica-compare} show that we produce much lower reconstruction errors than handcrafted renderers for UDFs. Visual comparisons in Fig.~\ref{fig:replica-visual-compare} show that our prior can recover geometry with sharper edges and much less artifacts. For objects that are only observed from one side, we are able to reconstruct it as a single surface, as illustrated in Fig.~\ref{fig:ablation-single-face}, which justifies the unsigned distance character.

\begin{figure}[t]
\centering
    \includegraphics[width=\linewidth]{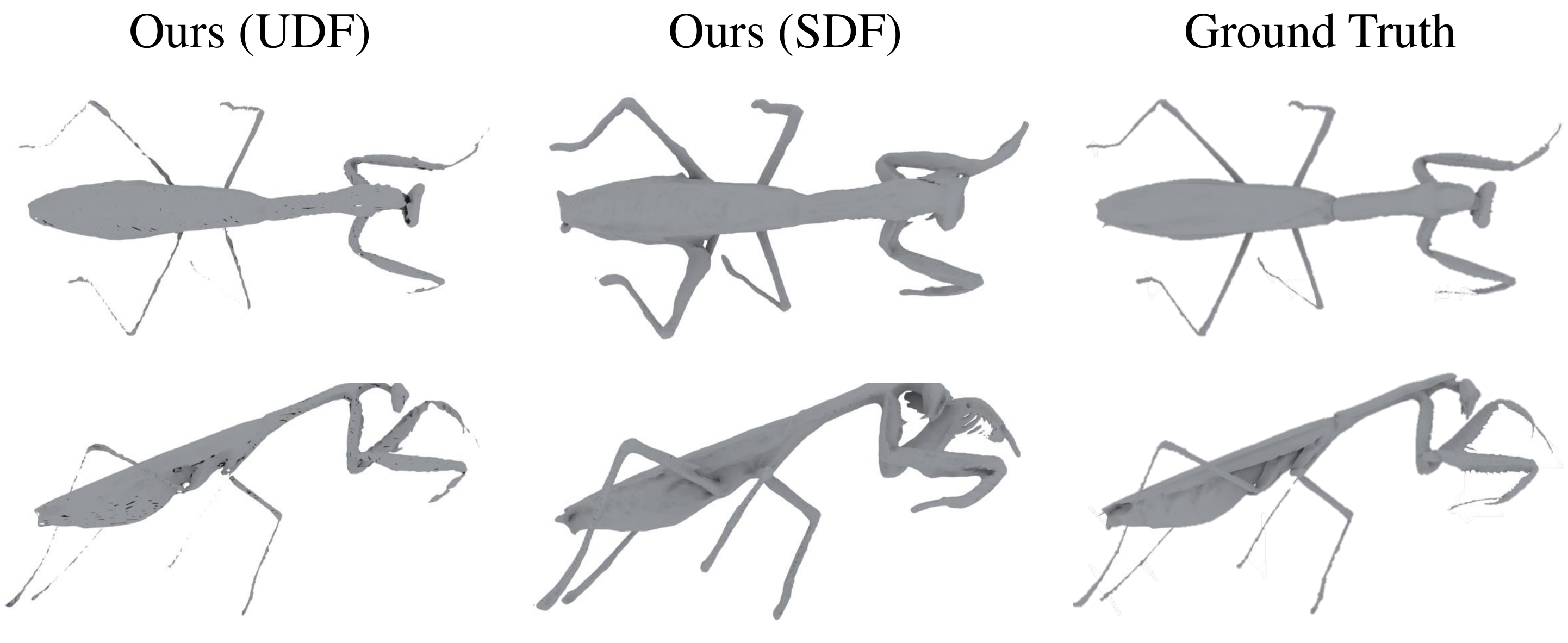}
    \caption{Comparison of UDF and SDF mesh extraction of our method on thin structures.}
  \label{fig:udf-sdf-extract}
\end{figure}
\subsubsection{Results on Insects} We further evaluate our method in reconstructing complex open surfaces using Insects dataset. Numerical comparisons (CD $\times 10^{-3}$) in Tab.~\ref{tab:insects-compare} demonstrate the superiority of our method. Visualization results in Fig.~\ref{fig:insects-visual-compare} show that existing UDF methods often produce incomplete surfaces with a mixture of single and double layer patches, while our method accurately reconstructs them. We categorize thin structures into two types: single-layer planes and thin cubes. Our method performs well on the former, such as the butterfly wings. The latter, however, are more challenging to reconstruct since their extremely small surface area may not be covered by the uneven and relatively sparse sampling during volume rendering. Furthermore, the inherent ambiguity of UDFs near the zero-level set limits the reconstruction quality of thin cubes under existing UDF mesh extraction algorithms, which is not the primary focus of our method. We provide a visual comparison between UDF and SDF reconstructions in Fig.~\ref{fig:udf-sdf-extract}, where the SDF result recovers the thin cube structures with inflated thickness, such as mantis legs, indicating that our learned UDFs indeed capture these details, but not in a complete manner. In future work, we plan to explore more advanced UDF extraction techniques~\cite{hou2023dcudf,stella2025high} to further improve the reconstruction quality of thin structures.

\subsubsection{Results on Real Scans} We further compare our method with NeUDF~\cite{liu2023neudf} and VRPrior~\cite{zhang2024learning} on real scans in Fig.~\ref{fig:teaser} and Fig.~\ref{fig:real-plant-fan-visual}. We shot $4$ video clips on $4$ scenes with thin surfaces, including McNuggets, eggshell, paper boat and potted flower. The comparisons in Fig.~\ref{fig:teaser} show that these challenging cases make NeUDF struggle to recover extremely thin surfaces like egg shells, resulting in incomplete and discontinuous surfaces. We also produce more accurate and smoother surfaces than VRPrior, for example, the ship's mast and sails, as well as the branches and flowers in the plants, as shown in Fig.~\ref{fig:real-plant-fan-visual}.

\begin{figure}[t]
\centering
    \includegraphics[width=\linewidth]{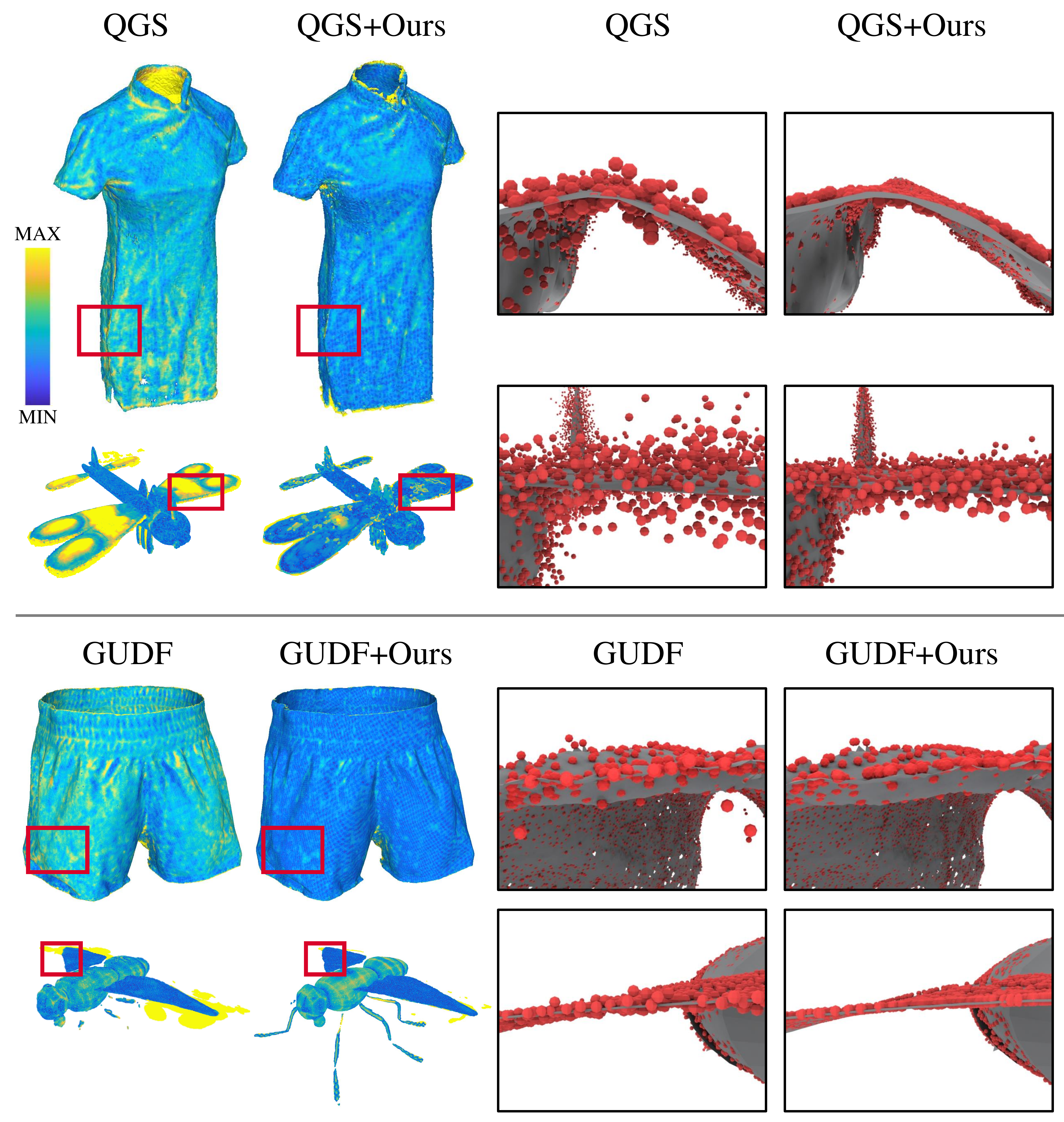}
    \caption{Visualization of our method integrated with different Gaussian Splatting methods. Left: Error maps of reconstructed meshes. Right: Gaussian centers around the ground truth surfaces. We learn much tighter Gaussian representations by leveraging VRPrior to impose a surface loss.}
  \label{fig:refine-gs-visual-errormap}
\end{figure}
\begin{table}[t]
  \centering
  \caption{\textcolor{red}{Comparison of CD($\downarrow$) for refining Gaussian Splatting using our VRP. We also report the average time (minutes) and memory (GB) consumption across three datasets.}}
  \label{tab:refine-gs}
  \resizebox{\linewidth}{!}{
  \begin{tabular}{l|ccccc}
    \toprule
     Datasets & DF3D & DTU & Insects & Time & Mem   \\
    \midrule
     \cmidrule{1-6}
     GaussianUDF\cite{li2025gaussianudf} & 1.60 & 0.68 & 4.34 & 48 & 1.2     \\  
     GaussianUDF+Ours                    & \textbf{1.45} & \textbf{0.65} & \textbf{3.54} & 59 & 2.1     \\  
    \cmidrule{1-6}
     QGS\cite{zhang2024quadratic}        & 4.46 & 0.54 & 3.57 & 21 & 2.0    \\  
     QGS+Ours                            & \textbf{2.97} & \textbf{0.53} & \textbf{3.13} & 33 & 3.1     \\  
  \bottomrule
    \end{tabular}}
\end{table}

\subsection{Extension to Gaussian Splatting Reconstruction}

Our pretrained volume rendering prior can serve as a general surface refiner that provides dense queries around the UDF zero-level set to optimize the Gaussian distributions toward the true surfaces, as described in Sec.~\ref{sec3.6}. We validate our efficiency by integrating our strategy into both GaussianUDF~\cite{li2025gaussianudf} and QGS~\cite{zhang2024quadratic}, and conducting experiments on DF3D, DTU, and Insects datasets. \textcolor{red}{We report the average CD as well as training time and memory consumption in Tab.~\ref{tab:refine-gs}. The resource costs during inference are identical to those of Gaussian Splatting, as the prior network is not used at inference.} Our method consistently improves performance across different variants, particularly on open surface datasets, where the reconstructed surfaces are more sensitive to Gaussian distributions. We further visualize the error maps of reconstructed meshes and the distributions of Gaussian centers in Fig.~\ref{fig:refine-gs-visual-errormap}. The results demonstrate that our method effectively optimizes the Gaussian distributions toward the true surfaces, leading to more accurate and faithful reconstructions.

\begin{figure}[t]
\centering
  \includegraphics[width=\linewidth]{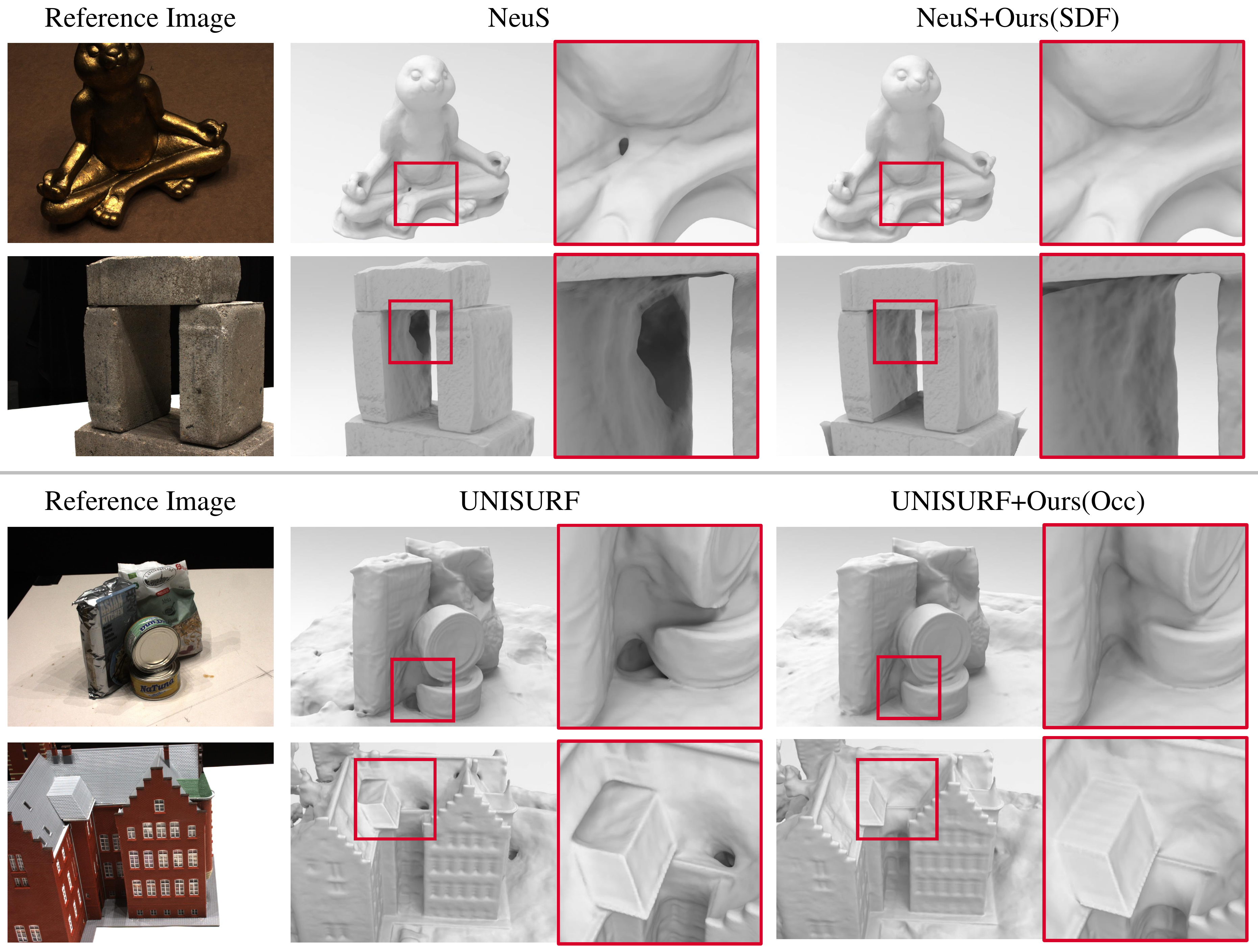}
    \caption{Effect of our volume rendering priors when we apply it to different neural representations, including signed distance functions (represented by NeuS~\cite{wang2021neus}) and occupancy (represented by UNISURF~\cite{oechsle2021unisurf}).}
  \label{fig:sdf-occ-visual}
\end{figure}

\subsection{Extension to Different Neural Representations}
Our volume rendering prior is not limited to unsigned distances. It can serve as a general module that transforms various neural implicit representations into rendering weights for alpha blending under different training settings. In this section, we evaluate our volume rendering priors when using SDF~\cite{park2019deepsdf} and occupancy~\cite{mescheder2019occupancy} as neural implicit representations to justify the generalizability of our framework.

\subsubsection{Signed Distance Function} To learn the weight function $f_{ws}$ for SDF inference, we first sample ground truth signed distances $f_{s_{gt}}^h$ from the ground truth meshes $\{S_h\}_{h=1}^H$ and then replace the UDF input $f_{u_{gt}}^h$ in Eq.~\eqref{eq:opaque} with SDF input $f_{s_{gt}}^h$. All other settings are same as those used for training $f_w$. During generalizing, we use our learned $f_{ws}$ instead of the differentiable renderer of NeuS~\cite{wang2021neus} to estimate the signed distance fields $f_s$. We report the reconstruction metrics on DTU dataset in Tab.~\ref{tab:sdf-occ-dtu}. The superiority in numerical results beyond NeuS demonstrate the effectiveness of our volume rendering priors when we apply the prior to signed distance functions. Visualization comparisons in Fig.~\ref{fig:sdf-occ-visual} show that our volume rendering prior corrects the reconstruction artifacts obtained with NeuS such as the holes and uneven surfaces.

\begin{table}[t]
  \centering
  \caption{Quantitative evaluations in terms of Chamfer Distance based on different neural representations in DTU dataset.}
  \label{tab:sdf-occ-dtu}
  \resizebox{\linewidth}{!}{
  \begin{tabular}{l|cccccccc|c}
    \toprule
    Methods & 24 & 37 & 40 & 55 & 63 & 97 & 110 & 122 & Mean \\
    \midrule
     NeuS\cite{wang2021neus}            & 1.07 & 1.21 & 0.73 & \textbf{0.40} & 1.20 & 1.16 & 1.59 & \textbf{0.51} & 0.98      \\  
     NeuS+Ours(SDF)                     & \textbf{0.94} & \textbf{1.05} & \textbf{0.56} & 0.42 & \textbf{1.14} & \textbf{0.98} & \textbf{1.33} & 0.52 & \textbf{0.87} \\
    \cmidrule{1-10}
     Unisurf\cite{oechsle2021unisurf}   & 1.73 & 1.42 & 1.59 & \textbf{0.45} & 1.30 & 1.13 & 1.43 & 0.71 & 1.22    \\
     Unisurf+Ours(Occ)                  & \textbf{1.25} & \textbf{1.17} & \textbf{1.11} & 0.52 & \textbf{1.22} & \textbf{0.97} & \textbf{1.25} & \textbf{0.62} & \textbf{1.01}    \\
  \bottomrule
    \end{tabular}}
\end{table}

\subsubsection{Occupancy} We adopt the similar way of learning SDFs and UDFs to learn the weight function $f_{wo}$ for occupancy inference. Note that the ground truth occupancies $f_{o_{gt}}^h$ are binary values, which will hinder the generalization ability of the priors if trained directly on such data. To address this issue, we use ground truth SDFs to approximate the real distribution of occupancy fields,
\begin{equation}
\label{eq:sdf2occ}
f_{o_{gt}}^h=\Phi(f_{s_{gt}}^h), \Phi(x)=\frac{1}{1+e^{20x}}.
\end{equation}
\noindent During inference, we accumulate the transformed occupancies $f_{wo}(f_o)$ as the weights for alpha blending, rather than directly accumulating occupancy values $f_o$ as in UNISURF~\cite{oechsle2021unisurf}. We also report numerical comparisons in Tab.~\ref{tab:sdf-occ-dtu}, where our results significantly outperform UNISURF, revealing the potential of our volume rendering priors in mitigating the bias from various differentiable renderers and neural implicit representations. Visual comparison in Fig.~\ref{fig:sdf-occ-visual} further shows that our volume rendering priors enable UNISURF to learn complete and smooth surfaces without holes.

\begin{figure}[t]
  \centering
  \includegraphics[width=\linewidth]{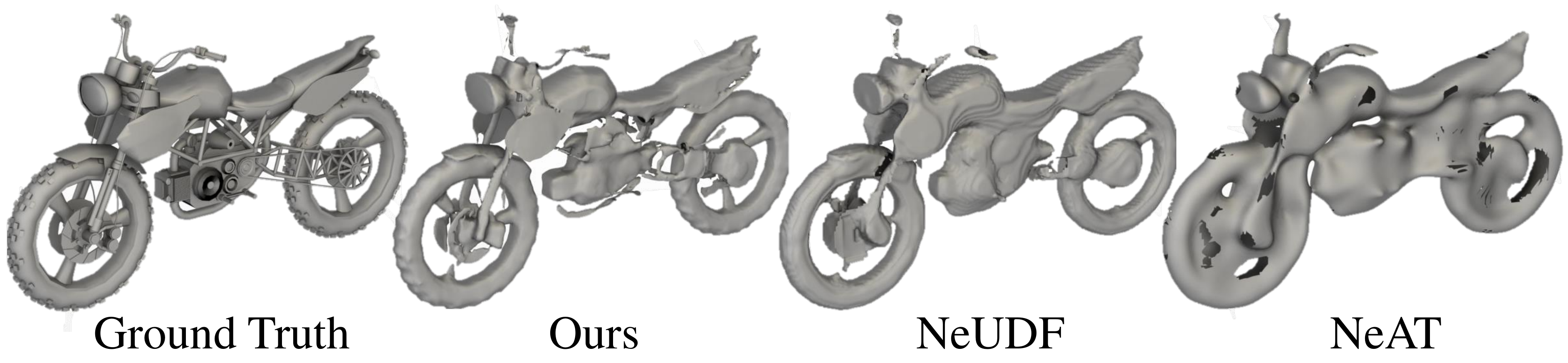}
  \vspace{-0.1cm}
  \caption{Comparison of the ability of overfitting single complex object using ground truth depth supervision.}
  \label{fig:ablation-shapenet-recovery}
\end{figure}
\begin{table}[t]
  \centering
  \caption{Ablation study on the choice of neighboring size. }
  \label{tab:neighboring-size}
  \resizebox{\linewidth}{!}{
  \begin{tabular}{cccccc|c|c}
    \toprule
    \multicolumn{6}{c|}{Neighboring Size} & \multicolumn{2}{c}{Datasets} \\
    \midrule
    1 & 10 & 20 & 30 & 40 & 50 & DTU\cite{jensen2014large} & DF3D\cite{zhu2020deepfashion3d} \\
    \cmidrule{1-8}
    \checkmark  &  &  &  &  &  & 1.01 & 2.26 \\
     & \checkmark &  &  &  &  & 0.87 & 2.21 \\
     &  &  & \checkmark &  &  & 0.85 & 1.71 \\
     &  &  &  &  & \checkmark & 0.89 & 1.99 \\
    \checkmark & \checkmark & \checkmark &  &  &  & 0.85 & 1.65  \\
     & \checkmark & \checkmark & \checkmark &  &  & \textbf{0.80} & \textbf{1.59}  \\
     &  & \checkmark & \checkmark & \checkmark &  & 0.83 & 1.73  \\
     &  &  & \checkmark & \checkmark & \checkmark & 0.89 & 1.97 \\
    \checkmark &  &  & \checkmark &  & \checkmark & 0.92 & 2.12  \\
  \bottomrule
    \end{tabular}}
\end{table}
\begin{table}[t]
  \centering
  \caption{Ablation study on the multi-resolution windows and hierarchical sampling strategies.}
  \label{tab:ablation-sampling}
  \resizebox{\linewidth}{!}{
  \begin{tabular}{l|c|c}
    \toprule
     & DTU\cite{jensen2014large} & DF3D\cite{zhu2020deepfashion3d} \\
    \midrule
    VRPrior~\cite{zhang2024learning} & 0.85 & 1.71 \\
    +Multi-resolution windows        & 0.82 & 1.67 \\
    +Point sampling priors          & 0.81 & 1.62 \\
    +Uniform sampling                & \textbf{0.80} & \textbf{1.59} \\
  \bottomrule
    \end{tabular}}
\end{table}
\begin{figure}[t]
  \centering
  \includegraphics[width=\linewidth]{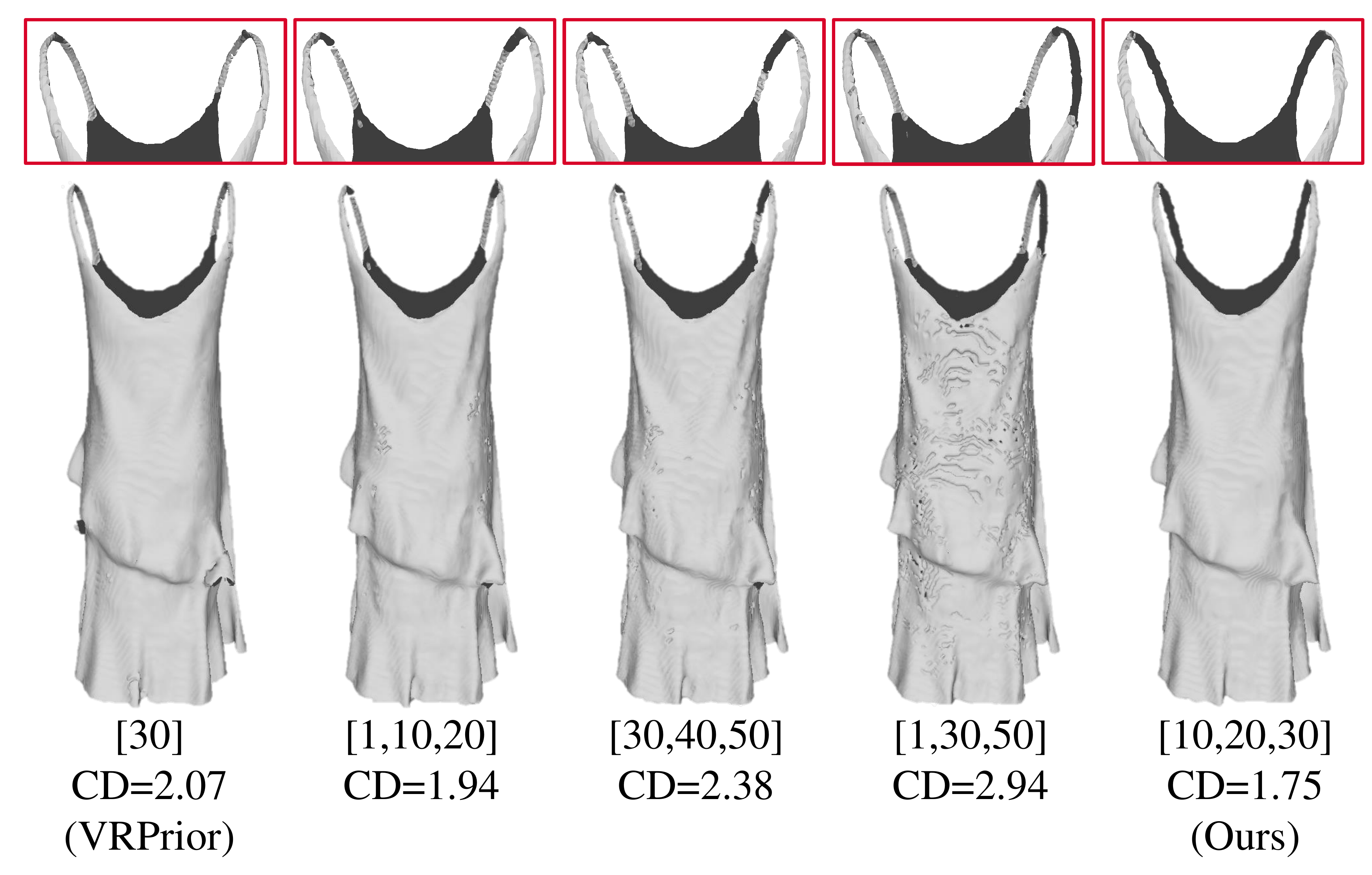}
  \vspace{-0.2cm}
  \caption{Ablation study on the neighboring size.}
  \label{fig:ablation-windowsize}
\end{figure}
\begin{figure}[t]
  \centering
  \includegraphics[width=.9\linewidth]{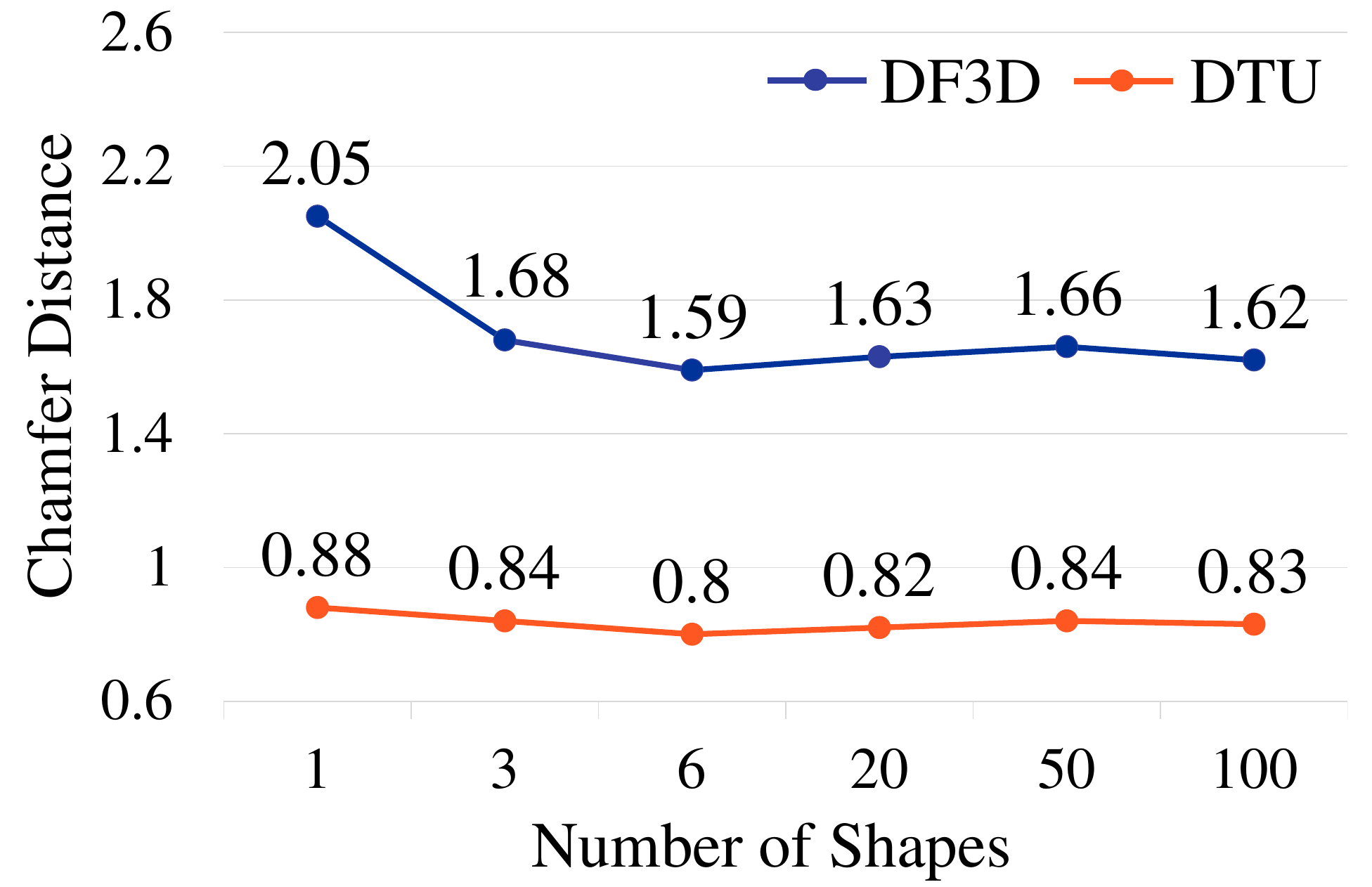}
  \vspace{-0.2cm}
  \caption{Ablation study on the number of shapes.}
  \label{fig:ablation-number-shapes}
\end{figure}
\begin{table}[t]
  \centering
  \caption{Ablation studies on prior variants. The numerical results are averaged across all objects in DF3D dataset.}
  \label{tab:ablation-variants}
  \resizebox{\linewidth}{!}{
  \begin{tabular}{l|c|c|c|c|c|c}
    \toprule
     Forms & Raw & \multicolumn{2}{|c|}{Inputs} & \multicolumn{2}{|c|}{Supervisions} & Inference  \\
    \midrule
     Variants & Ours & only UDF & UDF+Coor & RGB Sup. & RGBD Sup. & Finetune  \\
     \cmidrule{1-7}
     CD-L1 & \textbf{1.59} & 1.81 & 2.93 & $\times$ & 2.05 & $\times$ \\
  \bottomrule
    \end{tabular}}
\end{table}

\subsection{Ablation Studies and Analysis}

\subsubsection{Geometry Overfitting with Depth Supervision} We first report analysis on the capability of geometry reconstruction from depth images on a single shape, which highlights the performance of our prior over others without the influence of color modeling. We learn a UDF with our prior or other renderers from multiple depth images. Visual comparison in Fig.~\ref{fig:ablation-shapenet-recovery} shows that handcrafted renderers do not recover geometry detail even in an overfitting experiment, while our learned prior can recover more accurate geometry than others. 

\subsubsection{Neighboring Size} We report the effect of window size in our volume rendering prior on DF3D and DTU datasets in Tab.~\ref{tab:neighboring-size}, where we train different volume rendering priors with different single or multi-resolution window sizes including $\{1,10,20,30,40,50\}$. Compared to single window size, we find that multi-scale window sizes exhibit better generalization performance. This is because the multi-resolution setting enables the prior network to aggregate UDF variation information from different neighboring sizes, leading to a more accurate opaque density estimation. We also provide visualization results using different single or multiple window sizes, as shown in Fig.~\ref{fig:ablation-windowsize}. Small windows, such as $\{1,10,20\}$, make the prior sensitive to unsigned distance changes, while large-span windows, such as $\{1,30,50\}$, introduce ambiguity by blending information from vastly different resolutions. We find that window sizes covering $\{10,20,30\}$ queries work well in all of our experiments.

\subsubsection{Shapes for Learning Priors} We explore the effect of the number of shapes used for learning priors and report the generalization results on DF3D and DTU datasets, as represented in Fig.~\ref{fig:ablation-number-shapes}. The shapes are randomly selected from ShapeNet and DF3D datasets. The prior learned from a single shape exhibits a severe underfitting on DF3D, while more than six shapes do not bring further improvements. The reason is that we sample lots of rays from different view angles to provide adequate knowledge of transforming unsigned distances to densities, which covers almost all unobserved situations in volume rendering for the UDF inference during testing. Additionally, calculating GT UDFs from GT meshes for every sampled point is a time-consuming operation when learning priors, therefore we select six shapes for both efficiency and performance.

\subsubsection{Effectiveness of Sampling Strategies} We additionally report effects of our technical contributions over VRPrior~\cite{zhang2024learning}, as listed in Tab.~\ref{tab:ablation-sampling}. Building upon VRPrior, we progressively integrate each one of our modules to show the improvements of the reconstructed results on DTU and DeepFashion3D datasets. The first modification replaces the single window size in VRPrior with our multi-resolution windows. The second one introduces point sampling priors as an auxiliary network to refine the sampling near zero-level set during UDF inference. And the third one incorporates our uniform sampling strategy to promote an even distribution of upsampled points. The ablation results demonstrate the effectiveness of ours proposed modules. Although the numerical gains are limited by the NeuS backbone, our approach can be further enhanced through integration with other frameworks. For example, incorporating patch loss yields a notable performance improvement, as reported in Tab.~\ref{tab:dtu-patchloss}, while combining our strategy with Gaussian-based reconstruction methods leads to substantial additional improvements, as demonstrated in Tab.~\ref{tab:refine-gs}.

\subsubsection{Variations of training VRPrior} We discuss several possible variations in the training of our volume rendering prior. \textbf{Queries for Implicit Representations.} We justify the superiority of using additional sampling interval along the ray as queries. We try to remove the interval or replace the interval using other alternatives like coordinates. The degenerated results in the ``Inputs'' column in Tab.~\ref{tab:ablation-variants} indicate that the relative position represented by the interval generate better on unobserved scenes. \textbf{Supervisions for Learning Priors.} We further replace the supervisions of learning priors from depth images to RGB images or RGBD images, as reported in the ``Supervisions'' column in Tab.~\ref{tab:ablation-variants}. Training with only RGB supervisions does not converge while the RGBD supervision severely degenerates the performance due to the aliasing of the color net. This indicates that the color affects the generalization ability of the prior a lot and is not suitable for learning priors for UDF rendering. \textbf{Fine-tuning Priors.} Instead of using fixed parameters in the learned prior, we fine-tune the parameters of $f_w$ along with the optimization of $f_u$ during generalizing, as reported in the ``Inference'' column in Tab.~\ref{tab:ablation-variants}. We find that the optimization does not converge. This indicates that the prior has acquired sufficient generalization ability during training, requiring no further adjustments during testing.

%-----------------------------------------------------
%-----------------------------------------------------
\section{Conclusion}
We introduce volume rendering priors for UDF inference from multi-view images through neural rendering. We show that using data-driven manner to learn the prior can recover more accurate geometry than handcrafted equations in differentiable renderers. We successfully learn a prior from depth images from few shapes using our novel neural network and learning scheme, and robustly generalize the learned prior for UDFs inference from RGB images. We find that observing various unsigned distance variations during training and being 3D aware are the key to a prior with unbiasedness, robustness, and scalability. To further facilitate unbiased hierarchical sampling in UDF learning, we introduce point sampling priors and uniform sampling term as novel schemes. Extensive experiments and analysis on widely used benchmarks and real scans justify our claims and demonstrate both superiority over the state-of-the-art methods and strong generalization across various neural implicit representations.

% or


% \appendices

% % you can choose not to have a title for an appendix
% % if you want by leaving the argument blank
% \section{}

% % use section* for acknowledgment
% \ifCLASSOPTIONcompsoc
%   % The Computer Society usually uses the plural form
%   \section*{Acknowledgments}
% \else
%   % regular IEEE prefers the singular form
% \fi


\ifCLASSOPTIONcaptionsoff
  \newpage
\fi



%\bibliographystyle{IEEEtran}
% \end{thebibliography}
\bibliographystyle{IEEEtran}
\bibliography{references}


\begin{IEEEbiography}
[{\includegraphics[width=1in,height=1.25in,clip,keepaspectratio]{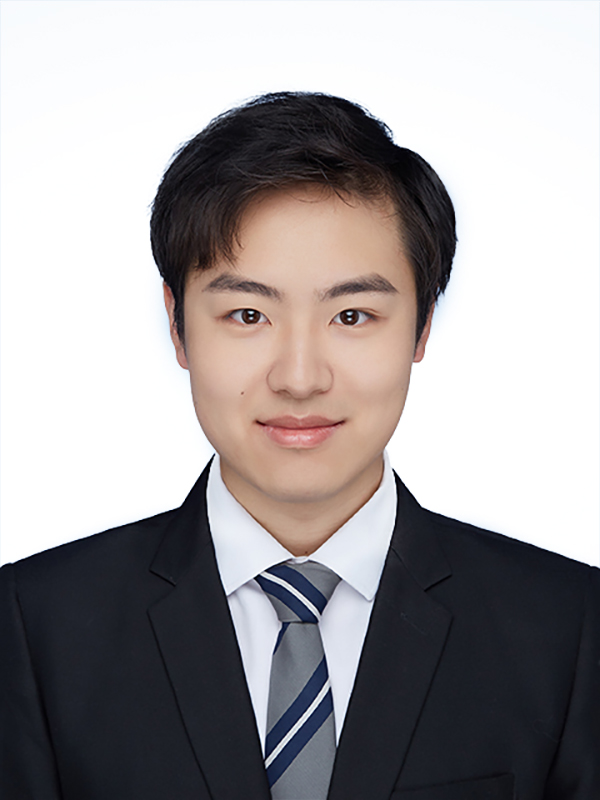}}]{Wenyuan Zhang}
    received the B.S. degree in School of Software from Tsinghua University, China, in 2021. He is currently the Ph.D. student with the School of Software, Tsinghua University. His research interests include novel view synthesis, multi-view reconstruction and 3D scene understanding.
\end{IEEEbiography}
\begin{IEEEbiography}[{\includegraphics[width=1in,height=1.25in,clip,keepaspectratio]{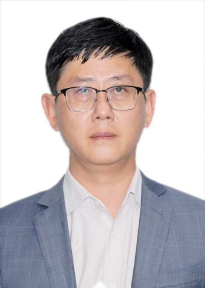}}]{Chunsheng Wang}
    received his B.S. degree from Northwest Normal University, in 2004. He is currently the deputy general manager in China Telecom Wanwei Information Technology Co., Ltd. His research interests cover spatial computing, AR/VR, 3D reconstruction.
\end{IEEEbiography}
\begin{IEEEbiography}[{\includegraphics[width=1in,height=1.25in,clip,keepaspectratio]{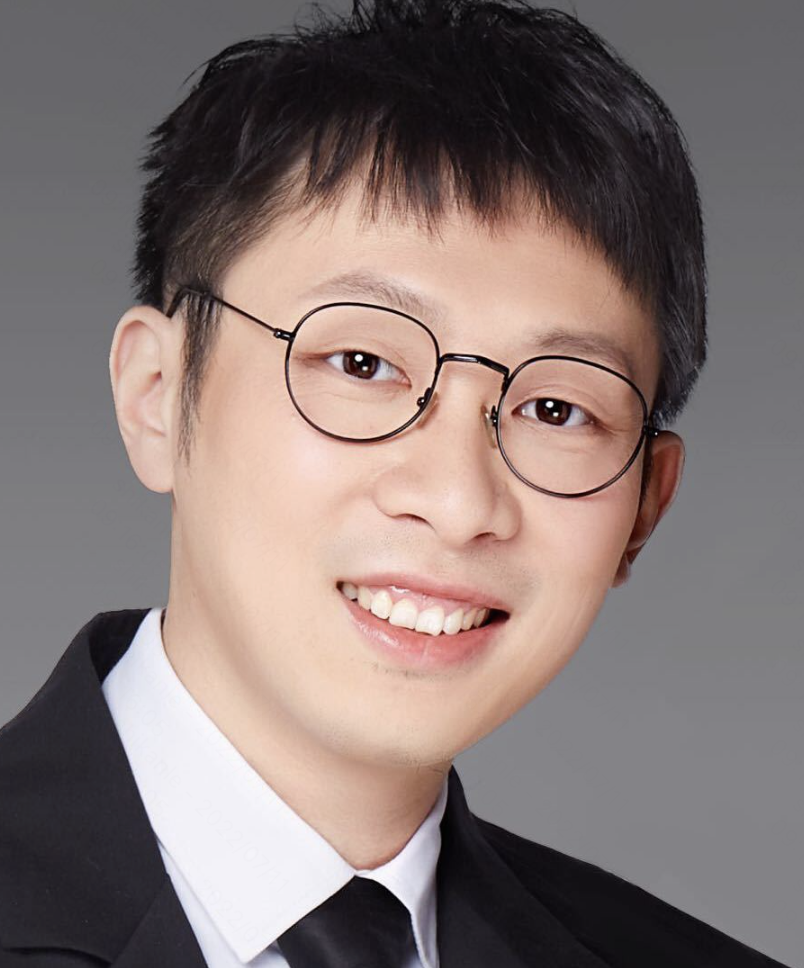}}]{Kanle Shi}
    received his B.S. degree in 2007, and Ph.D. degree in 2012, from Tsinghua University. From 2012 to 2013, he was a Post-Doctoral Researcher in INRIA, on computational geometry. From 2014 to 2019, he was an assistant/associate researcher at the School of Software, Tsinghua University. He is currently a researcher in Kuaishou Technology. His research interests cover computer graphics, geometry and machine learning.
\end{IEEEbiography}
\begin{IEEEbiography}[{\includegraphics[width=1in,height=1.25in,clip,keepaspectratio]{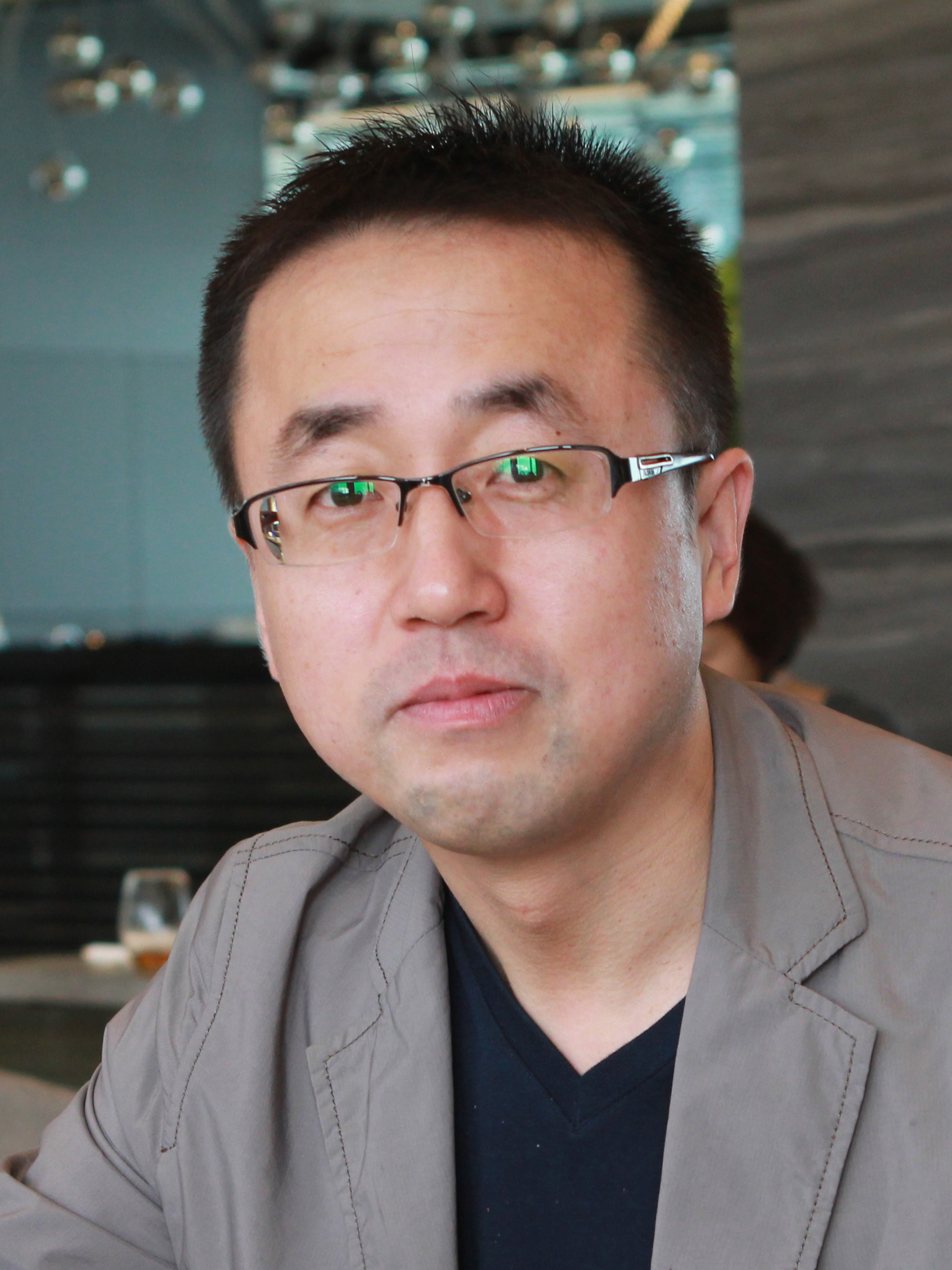}}]{Yu-Shen Liu}
    (M'18) received the B.S. degree in mathematics from Jilin University, China, in 2000, and the Ph.D. degree from the Department of Computer Science and Technology, Tsinghua University, Beijing, China, in 2006. From 2006 to 2009, he was a Post-Doctoral Researcher with Purdue University. He is currently an Associate Professor with the School of Software, Tsinghua University. His research interests include 3D computer vision, digital geometry processing, and 3D reconstruction.
\end{IEEEbiography}
%\newpage
\begin{IEEEbiography}[{\includegraphics[width=1in,height=1.25in,clip,keepaspectratio]{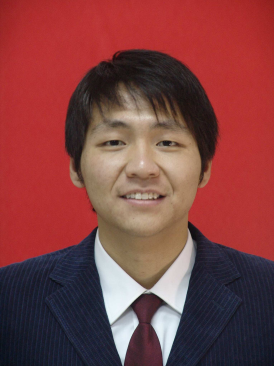}}]{Zhizhong Han}
    received the Ph.D. degree from Northwestern Polytechnical University, China, 2017. He was a Post-Doctoral Researcher with the Department of Computer Science, at the University of Maryland, College Park, USA. Currently, he is an Assistant Professor of Computer Science at Wayne State University, USA. His research interests include 3D computer vision, digital geometry processing and artificial intelligence.
\end{IEEEbiography}


%\vfill


\end{document}